\documentclass[runningheads]{llncs}
\usepackage{graphicx}

\usepackage[caption=false,font=normalsize,labelfont=sf,textfont=sf]{subfig}
\usepackage{amsmath,amsfonts}
\usepackage{array}
\usepackage{xcolor}
\usepackage{textcomp}
\usepackage{stfloats}
\usepackage{url}
\usepackage{verbatim}
\usepackage{graphicx}
\usepackage{cite}
\usepackage{graphicx}
\usepackage{placeins}
\usepackage{booktabs}
\usepackage{multirow}
\usepackage{tabularx}
\usepackage{algorithm}
\usepackage{algpseudocode}
\usepackage{scalerel}
\usepackage{tikz}
\usepackage{hyperref} 

\newcolumntype{C}{>{\centering\arraybackslash}X}
\newcolumntype{R}{>{\raggedleft\arraybackslash}X}
\newcolumntype{L}{>{\raggedright\arraybackslash}X}

\newfloat{algorithm}{t}{lop}

\algnewcommand\algorithmicsymbols{\textbf{Symbols:}}
\algnewcommand\ASymbols{\item[\algorithmicsymbols]}
\algnewcommand\algorithmicinput{\textbf{Input:}}
\algnewcommand\Input{\item[\algorithmicinput]}
\algnewcommand\algorithmicoutput{\textbf{Output:}}
\algnewcommand\Output{\item[\algorithmicoutput]}
\algnewcommand\algorithmicforeach{\textbf{for each}}

\algdef{S}[FOR]{ForEach}[1]{\algorithmicforeach\ #1\ \algorithmicdo}
\algdef{SE}{Unless}{EndUnless}[1]{\textbf{unless} \(\mbox{#1}\) \textbf{do}}{\textbf{end unless}}

\begin{document}
\title{\emph{Active Weighted Aging Ensemble} for Drifted Data Stream Classification}

\author{Michał Woźniak\orcidID{0000-0003-0146-4205} \and
Paweł Zyblewski\orcidID{0000-0002-4224-6709} \and
Paweł Ksieniewicz\orcidID{0000-0001-9578-8395}}
\authorrunning{M. Woźniak et al.}
%
\institute{Department of Systems and Computer Networks,\\
Wrocław University of Science and Technology,\\
Wybrzeże Wyspiańskiego 27, 50-370 Wrocław, Poland\\
\email{\{michal.wozniak;pawel.zyblewski;pawel.ksieniewicz\}@pwr.edu.pl}}
\maketitle              
\begin{abstract}
One of the significant problems of streaming data classification is the occurrence of concept drift, consisting of the change of probabilistic characteristics of the classification task. This phenomenon destabilizes the performance of the classification model and seriously degrades its quality. An appropriate strategy counteracting this phenomenon is required to adapt the classifier to the changing probabilistic characteristics. One of the significant problems in implementing such a solution is the access to data labels. It is usually costly, so to minimize the expenses related to this process, learning strategies based on semi-supervised learning are proposed, e.g., employing active learning methods indicating which of the incoming objects are valuable to be labeled for improving the classifier's performance. This paper proposes a novel chunk-based method for non-stationary data streams based on classifier ensemble learning and an active learning strategy considering a limited budget that can be successfully applied to any data stream classification algorithm. The proposed method has been evaluated through computer experiments using both real and generated data streams. The results confirm the high quality of the proposed algorithm over \emph{state-of-the-art} methods.

\end{abstract}

The paper focuses on constructing efficient data stream classifiers. Recently, most processed data is characterized by a large volume that comes to be processed as a data stream. This requires that the designed methods will take into account the streaming nature of the data and to select for this purpose appropriate processing so that the employed training algorithm can improve the recent classification model. On the other hand, it should be noted that traditional processing methods assume the stationarity of the classification task and thus do not take into account that probabilistic characteristics may change during the model lifetime. This phenomenon, called \emph{concept drift} \cite{Widmer:1996}, generally negatively affects the classification quality achieved by the model. Therefore, the proposed classifier training methods should also be able to cope with the elimination of the impact of \emph{concept drift} on the classifier quality.

The appearance of the \emph{concept drift} \cite{Widmer:1996} is common in many practical decision-making tasks, as fraudsters may change the content of the e-mail to get past spam filters. The occurrence of changes in the characteristics of a classification task is usually unpredictable. Still, we can identify factors that influence some problems, as the current pandemic situation related to the spread of the \textsc{covid}-\oldstylenums{19} virus very strongly influences consumer behavior. In the data classification task, the classifier aims to predict the class label $j$ ($j \in \mathcal{M}=\{1,…,M\}$, where $\mathcal{M}$ is a finite set of labels). The decision is made based on the attribute values of a given instance $x$ ($x=[x^{(1)},…,x^{(d)}]^T \in \mathcal{X}$, where $\mathcal{X}$ is a $d$-dimensional feature space), i.e., $\Psi\: :\: \mathcal{X} \to \mathcal{M}$. We assume that $x$ and $i$ are observed values of a pair of random variables $(\mathbf{X},\mathbf{J})$ \cite{Duda:2001}.

When we observe that the join distribution between two different time stamps varies, it means that \emph{concept drift} appears. The \emph{concept drift} impacts the mentioned probability distributions \cite{Gama:2013}, as either real or virtual \emph{concept drifts}. The first one means that changes will impact the shapes of decision boundaries, i.e., the \emph{posterior} probabilities $p(i|x)$ have been changed \cite{Schlimmer:1986,Widmer:1996}. The virtual drift does not alter the shape of decision boundaries, but it impacts the unconditional probability density functions \cite{Widmer:1993}. Olivera et al. \cite{Oliveira:2021} noted that although virtual \emph{concept drift} does not affect the change in decision boundaries and has not been the focus of much research, it is important to note that it can also affect the usefulness of learned decision boundaries by classifiers, e.g., for unrepresentative learning sets used to build a classifier. Therefore, in practical terms, it is not crucial what type of \emph{concept drift} occurs, at the end of the result in all scenarios, the current model needs to be altered.

Another taxonomy of concept drift is based on the rapidity of change. Mainly, we may distinguish (\emph{i}) sudden (or abrupt) \emph{concept drift}, when the new concept suddenly replaces the old concept; and (\emph{ii}) incremental concept drift when we may observe a steady progression from the old concept toward a new one such that at each time step the distance from the old concept increases and the distance to the new one decreases \cite{Webb:2018}. Minku et al. \cite{Minku:2010} proposed that probabilistic concept drift should also be considered, i.e., it occurs when there are two alternating concepts, such that initially, one concept dominates and over time, the other concept begins to dominate. However, many researchers, as Huang et al. \cite{Huang:2013} do not distinguish between these two types of drift. Gradual \emph{concept drift} should also be mentioned when the two concepts may occur with different intensities during the period of change between the old and the new concept. An interesting phenomenon is periodic changes, referred to as recurring \emph{concept drift} \cite{Sobolewski:2013} when previously occurring concepts return. This type of drift is typical for data streams associated with seasonal phenomena. In this case, we may observe a variant of recurring concept drift called a cyclical concept, when a specific number of concepts recur in a particular order \cite{Hoens:2012}.

Although many methods have been proposed that attempt to classify non-stationary data streams efficiently, there is still a need for new approaches that are the focus of intense research. Several methods dedicated to such classification can be found in the literature, including algorithms that continuously train a model of the classifier, the so-called online learners. Domingos \cite{Domingos:2003} formulated the following conditions that should be satisfied by such methods: (\emph{i}) each object must be processed only once during training; (\emph{ii}) the system should consume only a limited amount of memory and processing time, regardless of the execution time and the amount of data processed; (\emph{iii}) the training process can be stopped at any time, and its accuracy should not be lower than that of a classifier trained on batch data collected up to a given time. These methods are generally suitable for stationary data streams because all learning objects are equally valid, regardless of when they appeared. However, these methods could be applicable if we could control the process of forgetting objects from an outdated concept. Hence, among the methods used to classify non-stationary data streams, we can distinguish those that incorporate a forgetting mechanism. This approach is based on the assumption that the most relevant data have arrived recently, as they contain features of the current concept. However, their relevance decreases over time. Therefore, narrowing the scope of the data to those that have been read recently can help create a dataset that embodies the actual context. Three strategies are possible here: (\emph{i}) instance selection using a sliding window that cuts off older instances; (\emph{ii}) weighting the data based on relevance; and (\emph{iii}) using \emph{bagging} and \emph{boosting} algorithms that focus on misclassified instances.

For a sliding window, the main question is how to adjust the window size. On the one hand, a shorter window allows focusing on the emerging concept, although the data may not be as representative as in the case of a longer window. On the other hand, a wider window may result in a mixture of instances representing different concepts. Bifet and Gavalda \cite{BifetGavalda:2007} observed that although classifiers trained on wider windows are characterized by greater stability, they do not respond quickly enough to sudden concept drift. Examples of the relationship between window size, classifier accuracy and standard deviation are discussed in detail in \cite{Kurlej:2011}. Therefore, some advanced algorithms dynamically adjust the window size depending on the detected condition, e.g., \textsc{flora}\oldstylenums{2} \cite{Widmer:1996} and \textsc{adwin}\oldstylenums{2} \cite{Bifet:2007}. Then variances are compared using F-test and chunk size increases if the p-value is less than a predefined threshold. More advanced algorithms can even use multiple windows \cite{Lazarescu:2004}.

Another approach is to use so-called concept drift detectors, i.e., algorithms that can inform the classification system about changes occurring in the data stream distributions. A decision is made based on incoming information about new examples, i.e., labels or classifier performance are required to detect a \emph{concept drift}. Several approaches aim to detect drift from unlabeled data, but they are more suitable for detecting virtual concept drift \cite{Sobolewski:2013}. Additionally, many detectors can return both a signal that drift has been unambiguously detected and a warning level has been reached. It is a signal to start collecting new data to update or rebuild the model soon, i.e., when the explicit detection signal is returned. 

It is also important to realize that drift detection is a non-trivial task since the detection should be performed as soon as possible to replace the outdated model and thus minimize the reconstruction time. On the other hand, false alarms are unacceptable, as they will lead to misadaptation of the model and spending resources where there is no need to do so \cite{Gustafsson:2000}. 

Many drift detection methods have been proposed, including cumulative sum \cite{Page:1954}, which is a simple sequential analysis technique based on measuring the average value of the input data. The \emph{Exponentially Weighted Moving Average} \cite{Ross:2012} combines current and historical observations in a way that can quickly detect changes in the mean using aggregate chart statistics. Well-known \textsc{ddm} (\emph{Drift Detection Method}) \cite{Gama:2004} incrementally estimates the error of the classifier assuming convergence of the classifier training method.

\textsc{eddm} (\emph{Early Drift Detection Methods}) \cite{baena2006early} is an extension of \textsc{ddm}, where the window size selection procedure is based on the same heuristics. Interesting drift detectors based on Hoeffding and McDiarmid inequalities were proposed in \cite{Blanco:2015}. Also worth mentioning is \textsc{adwin} \cite{Bifet:2006}, which is based on an adaptive sliding window, being highly suitable for handling sudden drifts. While Nishida's algorithm \cite{Bifet:2007} assumes that the accuracy of the classifier for recent examples is the same as the overall accuracy since the beginning of learning if the target concept is stationary.

We also should mention combined models that combine the outputs of several detectors and then send a signal to a learning algorithm based on an ensemble decision. When one has an appropriate set of detectors and a good combination rule, the resulting ensemble detection can be expected to be more accurate and robust to noise. Maciel et al. propose a three-detector ensemble \cite{Maciel:2015}, based on different default detector combinations depending on the selected ensemble sensitivity. Du et al. \cite{du2014selective} developed the ensemble pruning technique to choose the most valuable component of the drift detector ensemble. Lapinski et al. \cite{Lapinski:2018} also studied several novel models of combined drift detectors.

The last approach to classifying non-stationary data streams is ensemble classifiers, which will be discussed in more detail in the next section. Here it is worth noting that the appropriate mechanisms must be developed to adapt the classification model to the changing probabilistic characteristics of the task. These can be divided into the following approaches \cite{Kuncheva:2004MCS}: (\emph{i}) \emph{dynamic combiners} -- base classifiers are trained before the model is run, and a combination rule (e.g., for weighted voting or aggregation by changing the weights associated with each classifier in the ensemble) is responsible for adapting to changes \cite{Littlestone:1994,Jacobs:1991}); (\emph{ii}) \emph{updating ensemble members} using recent training examples \cite{Oza:2001,Bifet:2010LevBag,Wang:2015}; (\emph{iii}) \emph{modifying ensemble line-up} \cite{Kolter:2003}.

An additional complication during the classification of non-stationary data streams is that the delivery of labeled data is required when we train the classifier under the changing probabilistic characteristics. This process can be difficult, primarily due to the frequent delay in delivering the correct classification. For example, the true label for the credit approval task for the client's score is not known until two years. The problem concerning the delay in accessing the label is significant. It should be noted that, especially in cases where the delay is substantial, the correct label may already be delivered when the evaluated object belongs to an outdated concept. Therefore, to the authors' best knowledge, there are currently no known solutions for this problem. However, the issue we want to address in this paper is the labeling cost. The labeling itself generally involves domain experts, so it is expensive. To reduce the expenses incurred for object labeling, it is possible to use methods that have their origin in semi-supervised learning methods, including those based on \emph{active learning} methods \cite{Greiner:2002}. Such an approach allows the selection of valuable learning instances according to their influence on the classifier's quality, which should be labeled. It thus positively influences the required budget needed for object labeling.

In a nutshell, the main contributions of this work are as follows:
\begin{itemize}
    \item The proposal of \textsc{awae} (\emph{Active Weighted Aging Ensemble for Drifted Data Stream Classification}) - a new chunk-base classifier ensemble for classification of non-stationary data streams.
    \item The proposition of new methods for weighting and aging (decoying) classifiers in the ensemble.
    \item The development of a method for rejuvenating base classifiers, i.e., increasing the decision impact of classifiers with above-average quality, despite the relatively long period of time that has passed since they were trained.
    \item The proposition of classifier pruning procedure based on an aggregate criterion that considers the influence of given base classifiers on ensemble diversity on the one hand and accuracy on the other.
    \item The proposal for an active learning method with limited labeling budget that can be applied to any chunk-based classifier ensemble.
    \item Carrying out the extensive experimental studies to evaluate the quality of the proposed method, emphasizing the effect of individual parameters on learning quality and comparing classification performance with selected \emph{state-of-the-art} methods.  
\end{itemize}

The structure of this article is as follows. The following section will briefly present works related to classifier ensemble for the non-stationary data stream and active learning. Then \textsc{awae} will be discussed, and the  active learning pool-based sampling technique with limited labeling budget \textsc{bals} for data stream classifiers will be presented. The following section contains the experimental results and answers to the researcher questions posed at the beginning of this section. The last part summarizes the work and indicates directions for possible further research.

\section{Related works}
\label{sec:relworks}

The data stream can be split into into small data blocks called chunks. Therefore learning from these portions of data are called batch-based or chunk-based learning \cite{Krawczyk:2017}. Choosing the proper size of the chunk is crucial because it may significantly affect the classification \cite{Junsawang:2019}. Although there are methods mentioned earlier to adjust the size of the data chunk to changing distributions, most chunk-based classification methods assume that the size of the data chunk is priorly set and remains unchanged during the data processing. Instead of chunk-based learning, the algorithm can learn incrementally (online), i.e., training examples arrive one by one at a given time and are not kept in memory. 

When processing a non-stationary data stream, we can rely on a drift detector to point moments when data distribution has changed and take appropriate actions. Alternatively, we may use inherent adaptation capabilities of models. One class of such models is classifier ensemble \cite{Kuncheva:2014}.

\subsection{Classifier ensemble for non-stationary data stream}

Firstly, it is worth mentioning work by Krawczyk et al.~\cite{Krawczyk:2017} where a comprehensive overview of classifier ensemble learning from data streams is presented. Due to processing data streams, we may employ online and chunk-based ensemble learning. Within the first group of methods, we should first mention \emph{Online Bagging} proposed by Oza and Russel~\cite{Oza:2001}, which is inspired by \emph{bagging}, whereby when a new object arrives, it is used to train each base classifier. The number of recent object presentations for each base classifier is determined by $Poisson(1)$ distribution. The proposed method was later developed by Lee and Clyde~\cite{Lee:2004} in the \emph{Bayesian Online Bagging} algorithm. In contrast, Bifet et al. proposed \emph{Adaptive-Size Hoeffding Trees} algorithm. It ensures ensemble diversity by learning Hoeffding trees of different sizes. \emph{Leverage Bagging}~\cite{Bifet:2010LevBag} combines the simplicity of \emph{bagging} with adding more randomization to the input and output of the classifiers by using $Poisson(\lambda)$ distribution, where the user can determine $\lambda$. \emph{Leverage Bagging} also uses random output codes and allows training new base classifiers when concept drift is detected using \textsc{adwin}.

Oza and Russell~\cite{Oza:2001} also authored \emph{Online Boosting}. It employs a sequence of base models trained during the \emph{boosting} procedure. When a new learning example is received, each base classifier is updated multiple times according to the $Poisson(\lambda)$ distribution. For the first classifier in the sequence $\lambda=1$ and in the case of misclassification, the parameter $\lambda$ is increased for the next base classifier, and in the case of correct classification, it is decreased accordingly. 

Several important modifications of the above methods are also worth mentioning. Santos et al.~\cite{SantosADOB:2012} proposed \emph{Adaptable Diversity-based Online Boosting} (\textsc{adob}), which can accelerate the update of base classifiers after concept drifts by making the $\lambda$ parameter dependent on the prediction quality of base classifiers. Based on this approach, \emph{Accuracy Weighted Diversity-based Online Boosting} (\textsc{awdob}) \cite{Baidari:2020} employs weighted voting according to previous base classifier evaluations results. Baros et al. developed \emph{Boosting-like Online Learning Ensemble} (\textsc{bole})~\cite{Barros:2016}, which used a modification of the \textsc{adob} algorithm involving weakening the requirements to allow the base classifier to vote and use different concept drift detector. Also \emph{Ultra Fast Forest Tree} (\textsc{ufft})~\cite{GamaUFFT:2004} is worth mentioning. It uses an ensemble of online training binary Hoeffding trees. Lan et al.~\cite{Lan:2009} proposed \emph{Ensemble of Online Extreme Learning Machines} being a simple aggregation of randomized neural networks trained online. Shan et al.~\cite{Shan:2018} proposed an online active learning ensemble that composed a long-term stable classifier and multiple dynamic classifiers. This method also used an active learning strategy to select instances to label.

Let us move on to the second group of chunk-based stream processing methods to build classifier ensembles. \emph{Streaming Ensemble Algorithm} (\textsc{sea})~\cite{Street:2001} is the classifier ensemble with changing lineup, where the individual classifiers are trained on the following data chunks. The base classifiers with the lowest accuracy are removed from the ensemble to keep the model up-to-date. Wang et al. proposed \emph{Accuracy Weighted Ensembles} (\textsc{awe})~\cite{Wozniak:2013} employing the weighted voting rules, where weights depend on the accuracy obtained on the testing data. Brzezinski and Stefanowski proposed \emph{Accuracy Updated Ensemble} (\textsc{aue}), extending \textsc{awe} by using online classifiers and updating them according to the current distribution~\cite{Brzezinski:2011}. Shan et al.~\cite{Shan:2018} developed an online classifier ensemble consisting of a so-called \emph{stable classifier} and multiple updating base classifiers to better react to different concept drifts. This approach also employed active learning to minimize the required number of labeled instances.

Cano and Krawczyk proposed \emph{Kappa Updated Ensemble} (\textsc{kae})~\cite{Cano:2020} that combines online and block-based approaches. \textsc{kae} uses Kappa statistic for dynamic weighing and selection of base classifiers. It is also worth mentioning work by Cohen and Straus~\cite{Cohen:2003} where the problem of maintaining time-decaying was formulated and analyzed, and statistics of a data stream. Liu et al.~\cite{Liu:2017} proposed dividing data chunks in case of drift occurrence. Lu et al.~\cite{Lu:2020} compared two chunks based on classifier predictions variance calculated using the original method called \emph{subunderbagging}.

Bifet et al.~\cite{Bifet:2007} introduced a method for handling concept drift with varying chunk sizes. Each incoming chunk is divided into two parts: older and new. Empirical means of data in each subchunk are compared using Hoeffding bound. If the difference between two means exceeds the threshold defined by confidence value, then data in the older window is qualified as out of date and is dropped. Later window with data for current concept grows, until next drift is detected and data is split again. This approach allows for detecting drift inside the chunk.

\subsection{Active learning methods for data stream classification}

Due to the high cost of data labeling, active learning methods are gaining more and more popularity~\cite{Settles:2012}, including data stream classification~\cite{Attenberg:2011}.

\v{Z}liobait\.{e} et al. \cite{Zliobaite:2014} discussed the theoretical framework for predictive model learning using active learning approach and described tree labeling strategies. Kurlej and Wozniak \cite{Kurlej:2012} proposed the active learning data stream classification methods based on minimal distance classifiers. The decision about a~given instance labeling depends on the distance between an example and the current decision boundary. Nguyen et al. \cite{Nguyen:2013} developed an incremental algorithm {\textsc{csl}-stream}, that performs clustering and classification at the same time. Zgraja et al.~\cite{Zgraja:2018} developed \textsc{alcc} (\emph{Active Learning by Clustering for Drifted Data Stream Classification}) algorithm which employed query by clustering into new classifier training.

Mohamad~\cite{Mohamad:2018} proposed a similar data stream classifier, which combines uncertainty and density-based querying criteria. Bouguelia et al.~\cite{Bouguelia:2016} proposed a new query strategy based on instance weighting, but the time complexity of the weigh calculation was relatively high. Korycki and Krawczyk~\cite{Korycki:2018} applied an active learning strategy to classify data streams and combine it with the self-labeling approach.

Ksieniewicz et al.~\cite{Ksieniewicz:2019} focused on active learning strategy for neural network classifiers. The authors implemented the forgetting mechanism using the \emph{catastrophic forgetting} phenomenon. Shan et al.~\cite{Shan:2018} employed a mixed strategy (based on active learning uncertainty sampling and random sampling) to select incoming objects to label. This approach was inspired by~\cite{Xu:2016} where the proposed algorithm reacts to \emph{concept drift} by using a mixed strategy to choose the instances to be labeled.

\section{Active Weighted Aging Ensemble}
\label{sec:wae}

\subsection{Preliminaries}

Let’s $\Pi$ denotes a pool of $L$ base classifiers $\Pi=\{\Psi_1, \Psi_2,\Psi_k ..., \Psi_L\}$ to be used by the combined classifier $\hat{\Psi}$. 

In this study, we employ weighted voting, where weights are assigned to each base classifier, i.e.,

\begin{equation}
\label{eq.3}
\hat{\Psi}(x)=\mathop{\arg\max}\limits_{i \in \mathcal{M}} \mathop{\sum}\limits_{k=1}^L \left[\Psi_k(x)=i\right]w(\Psi_k) \end{equation}

\noindent where $w_k$ is the weight that is assigned to the $k$th individual and $[\;]$ denotes Iverson's bracket.

\subsection{Algorithm decsription}

In this section, we will discuss a novel \textsc{awae} algorithm (\emph{Active Weighted Aging Ensemble}), which can adapt the model to changes caused by the appearance of \emph{concept drit}. On the other hand, it reduces the cost of labeling the data in each data chunk by using an active learning strategy with a maximum labeling budget assumed. It is worth noting here that the proposed algorithm is inspired by previous works of the team related to the \textsc{wae} algorithm~\cite{Wozniak:2013}, among others. However, the presented version has a number of modifications in terms of reducing the label demand and proposes new mechanisms for calculating weights for base classifiers and ensemble pruning. 

We will first briefly introduce the most critical components of the \textsc{awae} algorithm, propose their implementation, and then present a run-time analysis of the proposed method.

Instead of drift detection, \textsc{awae} tries to construct a self-adapting classifier ensemble that can adapt to the changes in a data stream. We assume that the classified data stream is given in the form of data chunks denoted as ${DS}_{k}$, where $k$ is the chunk index. The concept drift could appear in the incoming data chunks. Therefore based on each chunk, one individual is trained and we check if it could form a valuable ensemble with the previously trained models.

If the pool size of $L$ classifiers is exceeded, then the least valuable classifier must be removed from the pool (we choose $L$ out of $L+1$ individuals). This choice (ensemble pruning) is made based on a proposed criterion that is either the chosen diversity measure (in the case of \emph{pre\_pruning}, i.e., before the weights are computed) or a linear combination of diversity measure and accuracy (in the case of \emph{post\_pruning} after the weights of the base classifiers are calculated). Three procedures are used to calculate the weights. The \emph{weight calculating procedure} calculates weights according to the importance of a given classifier for ensemble quality. The \emph{rejuvenating procedure} artificially reduces the residence counter of classifiers with a high impact on the quality of the ensemble classifier, i.e., it weakens the forgetting effect associated with the selected classifiers. In contrast, \emph{aging procedure} reduces the weights of the base classifiers depending on the value of the residence index. The detailed description of the \textsc{awae} is presented in Alg. \ref{alg:awae}.

\begin{algorithm}[!htb]
\caption{\emph{Active Weighted Aging Ensemble} algorithm}
\label{alg:awae}
\begin{algorithmic}[1]
\Input 
    \Statex data stream $DS=\{DS_1, DS_2, ...\}$,
    \Statex data chunk size,
    
    \Statex $train$ --- classifier training function using active learning stategy;
    \Statex $L$ -- maximal ensemble size;
    \Statex $pruning$ -- pruning procedure;
    \Statex $pre\_pruning$ -- pre\_pruning procedure on/off;
    \Statex $post\_pruning$ -- post\_pruning procedure on/off;
    \Statex $criterion\_p$ -- pruning criterion;
    \Statex $diversity$ -- diversity measure;
    \Statex $weighting$ - weight calculating procedure;
    \Statex $aging$ -- aging procedure;
    \Statex $reindex$ - reindexing of classifier identifiers starting from 1;
    \Statex $reju$ -- rejuvenating procedure.
    \vspace{.25em}
    
    \State $k:=0$
    \State $\Pi=\emptyset$
    \While{new data chunk $DS_k$}
            \State $\Psi_k\leftarrow train(DS_k)$
            \State $\Pi := \Pi\;\cup\;\{\Psi_{k}\}$ 
        \If{$pre\_pruning$}
            \State $\Pi \leftarrow pruning(\Pi, L, diversity)$
        \EndIf
        \State $reindex(\Pi)$
        \State $w := 0$
        
        \State $weighting(\Pi,DS_k)$
        \State $reju(\Pi,DS_k)$
        \State $aging(\Pi,DS_k)$
        \For{$j:=1$ \textbf{to} $|\Pi|$}
        \If{$w(\Psi_j)==0$}
                \State $\Pi=\Pi\setminus\{\Psi_j\}$
            \EndIf
            \State $w:=w+w(\Psi_j)$
        \EndFor
        \State $reindex(\Pi)$
        \If{$post\_pruning$}
            \State $\Pi \leftarrow pruning(\Pi, L, criterion_p)$
            \State $reindex(\Pi)$
            \For{$j:=1$ \textbf{to} $|\Pi|$}
                \State $w:=w+w(\Psi_j)$
            \EndFor
        \EndIf
        
        \For{$j:=1$ \textbf{to} $|\Pi|$}
            \State $w(\Psi_j) := \frac{w(\Psi_j)}{w}$
        \EndFor
        \State $k:=k+1$
    \EndWhile
\end{algorithmic}
\end{algorithm}

\subsection{Ensemble pruning}
\label{sel-criterion}

\textsc{awae} has built-in mechanisms to ensure a maximum ensemble size. If this is exceeded, a pool of $L$ base classifiers with the best evaluation function value is selected. Pruning is possible either as soon as a classifier is added to the ensemble or only after determining its weight used by the combination rule. In the first case, pruning is performed if the variable $pre\_pruning==1$ and the selected \emph{diversity measure} ($Generalized Diversity$ proposed by Partridge and Krzanowski \cite{Partridge:1997} was selected in the experiments) is used for evaluation~\cite{Kuncheva:2014}. When post pruning is selected (variable $post\_pruning==1$), the following criterion is used 

\begin{equation}
\label{criterion}
criterion_p(\Pi)=\alpha P_a(\Pi)+(1-\alpha)diversity(\Pi)
\end{equation}
where $P_a(\Pi)$ stands for the accuracy of the classifier ensemble, $diversity(\Pi)$ denotes its diversity, while $\alpha \in[0,1]$ is user-defined parameter used for the linear combination of accuracy and diversity. Both metrics are calculated on the basis of incoming data chunk.

\subsection{Weight calculation}
We propose the following methods of weight calculation: \\
\noindent \textbf{The same weights for each classifier} in the pool, i.e., majority vote is use as the combination rule
\begin{equation}
w(\Psi_i)=\frac{1}{|\Pi|}
\end{equation}

\noindent \textbf{Kuncheva's weights} - suggested by Kuncheva  \cite{Kuncheva:2014}
\begin{equation}
w(\Psi_i)=\frac{P_a(\Psi_i)}{1-P_a(\Psi_i)}
\end{equation}

\noindent \textbf{Weights proportional to accuracy related to the whole ensemble accuracy} 

\begin{equation}
w(\Psi_i) =\frac{P_a(\Psi_i)}{P_a(\Pi)}
\end{equation}
if $w(\Psi_i)<\theta$ then $w(\Psi_i)=0$, where $\theta \in [0,1]$ is the parameter which responsible for the removing less important classifiers.

\noindent \textbf{Weights proportional to accuracy related to the whole ensemble accuracy using bell curve}
\begin{equation}
w(\Psi_i)= \frac{1}{2\pi}exp\frac{(P_a(\Pi)-(P_a(\Pi) - \Psi_i))}{2} 
\end{equation}
if $w(\Psi_i)<\theta$ then $w(\Psi_i)=0$, where $\theta \in [0,1]$ as previously stands for the parameter responsible for the removing less important classifiers.

\subsection{Aging}
We propose the following aging methods:

\noindent \textbf{Weight aging proportional to classifier accuracy}

\begin{equation}
w(\Psi_i)=\frac{P_a(\Psi_i)}{\sqrt{itter(\Psi_i)}}
\end{equation}

\noindent where $itter(\Psi)$ stands for the residence counter of the classifier $\Psi$ in $\Pi)$, i.e., how many iterations have elapsed since a given base classifier was trained.

\noindent \textbf{Constant aging}
\begin{equation}
w(\Psi_i) = w(\Psi_i)=w(\Psi_i)-\delta 
\end{equation}
if $w(\Psi_i)<\theta$ then $w(\Psi_i)=0$, where $\theta \in [0,1]$ as previously stands for the parameter responsible for the removing less important (old enough) classifiers and $\delta$ is user-defined parameter responsible for the aging rate.

\noindent\textbf{Gaussian aging}

\begin{equation}
w(\Psi_i) = \frac{1}{2\pi}exp\frac{itter(\Psi_i)\xi}{2}
\end{equation}
if $w(\Psi_i)<\theta$ then $w(\Psi_i)=0$, where $\theta \in [0,1]$ as previously stands for the parameter responsible for the removing less important (old enough) classifiers and $\xi$ is user-defined parameter responsible for the aging rate.

\subsection{Rejuvenating}

We propose to rejuvenate an individual classifier if it has a big impact on the classifier ensemble, i.e., if its weight is bigger than average weight of base classifiers. Then the residence counter of a base classifier in the ensemble ($itter$) is decreased. The idea is presented in Alg.~\ref{alg:rejuv}, where $[ ]$ stands for \emph{entier}.




\begin{algorithm}[!htb]
\caption{Rejuvenating}
\label{alg:rejuv}
\begin{algorithmic}[1]
\Input 
    \Statex $r\_p$ -- power of rejuvenating ($r\_p>1$),
    \Statex $\{w(\Psi_1),w(\Psi_2)...\}$ -- weights  assigned to the individuals in $\Pi$.
    \vspace{.25em}
    
    \State $w:=0$
    \For{$j:=1$ \textbf{to} $|\Pi|$}
        \State $w:=w+w(\Psi_j)$
    \EndFor
    \State $w:=\frac{w}{\Pi}$
    \For{$j:=1$ \textbf{to} $|\Pi|$}
        \If{$w(\Psi_j)>w$}
            \State $itter(\Psi_j):=itter(\Psi_j)-$ 
            \Statex \hspace{3em}$max(1,[r\_p*(w(\Psi_j)])$
        \EndIf
    \EndFor
\end{algorithmic}
\end{algorithm}

\subsection{Budget Active Labeling Strategy}

The training algorithm $train$ used by \textsc{awae} algorithm also takes into account the possibility of limited label access, in which the \emph{Budget Active Labeling Strategy} (\textsc{bals}) algorithm~\cite{zyblewski2020combination} -- based on the random budget and active learning paradigm -- is employed. The decision about the object labeling in each consecutive data chunk depends on two parameters:

\begin{itemize}
    \item threshold $t$ -- which is responsible for choosing the "interesting" examples, i.e., if support function related with the decision is lower than a given threshold the object seems to be interesting and algorithm is asking for its label.
    \item budget $b$ -- which defines the percent of instances in each data chunk, for which a label will be randomly obtained.
\end{itemize}

The addition of the randomly chosen budget to the instances selected using the threshold-based active learning method aims at increasing the generalization ability of a classifier while reducing the possibility of overfitting. The exact procedure of the processing performed by the \textsc{bals} algorithm is presented in the Alg.~\ref{alg:bals}, and below is a brief description of the functions used in it:

\begin{itemize}
    \item $update\_classifier()$ -- Updates the classification model using the labeled instances from a given data chunk, or with whole data chunk in case of the first iteration.
    \item $active\_learning()$ -- Selects instances for labeling from a given data chunk based on the support function threshold, which can be interpreted as a distance from the decision boundary in the case of binary classification problems. 
    \item $random\_budget()$ -- Selects a random percent of samples from a given data chunk for labeling.
    \item $get\_labels()$ -- Obtains the labels for the selected instances.
\end{itemize}

\begin{algorithm}[!htb]
\caption{Budget Active Labeling Strategy}
\label{alg:bals}
\begin{algorithmic}[1]
\Input 
\Statex input data stream,
\Statex data chunk size
\Statex $\Psi$ -- classification algorithm,
\Statex $t$ -- threshold value,
\Statex $b$ -- budget value.
    \vspace{.25em}
    
    \State $k:=0$
    \While{new data chunk $\mathcal{DS}_k$}
        \If{$k==0$}
            \State $\Psi \gets update\_classifier(\Psi, \mathcal{DS}_k)$ 
        \Else
          \State $\mathcal{X}_k = active\_learning(t, \mathcal{DS}_k)$ 
          \State $\mathcal{X}_k \gets random\_budget(b, \mathcal{DS}_k)$ 
            \State $\mathcal{LS}_k = get\_labels(\mathcal{X}_k)$ 
            \State $\Psi \gets update\_classifier(\Psi, \mathcal{LS}_k)$ 
        \EndIf
    \State $k:=k+1$
    \EndWhile
\end{algorithmic}
\end{algorithm}

\subsection{Run-time analysis}

The \textsc{awae} algorithm can be decomposed into a few stages. First, for each data chunk, a new base classifier is trained. Training time depends on the type of classification algorithm but is constant for each instance. This results in computational complexity of $O(\rvert DS_i \rvert)$ for each new classifier training. Then, the rejuvenation process is performed for each base model, with the computational complexity of $O(\rvert \Pi \rvert)$. If the maximal ensemble size $L$ is exceeded, the pruning process is performed. By limiting the possible combinations of base classifiers in the pool only to those containing $\rvert \Pi \rvert-1$ elements, the computational complexity of pruning is $O(L)$. Next, the weight calculation and aging are performed for each base model with a complexity of $O(\rvert \Pi \rvert)$ and finally, all weights are updated with the same computational complexity $(O(\rvert \Pi \rvert|))$.

The \emph{active learning} \textsc{bals} algorithm is composed of two stages. First, it calculates each sample's distance from the decision boundary (which is an absolute difference of obtained support and $.5$), which has the complexity of $O(\rvert \mathcal{DS}_i \rvert)$. Then the objects are sorted according to the obtained distance and only those whose distance values do not exceed the set threshold $t$ are used. This operation has the computational complexity of $O(\rvert \mathcal{DS}_i \rvert$ $log$ $\rvert \mathcal{DS}_i \rvert)$. In the second stage, \textsc{bals} uses a simple sampling without replacement in order to choose the random budget $b$ of instances from each data chunk $\mathcal{DK}_i$. This operation has the computational complexity of $O(b$ $log$ $b)$.

\section{Experimental study}

In this section, we will describe the details of a conducted experimental study that can assess the usefulness of \textsc{awae}



The experiments are designed to answer the following research questions:
\begin{itemize}
    \item[RQ1.] What is the best parameter setting for \textsc{awae} and how it impact the behaviour of the proposed algorithm as 
    its ability to classify data streams with concept drift?
    \item[RQ2.] How flexible is \textsc{awae} to be used with the different classifiers?
    \item[RQ3.] How does \textsc{awae} behave when there is limited access to the labels? 
    \item[RQ4.] How does the \textsc{awae} compare to the \emph{state-of-the-art} algorithms,
     explicitly designed for the drifting data streams classification tasks?
    \item[RQ5.] How do the selected \emph{state-of-the-art} algorithms behave when using the proposed \textsc{bals} active learning strategy?
\end{itemize}

\subsection{Set-up}


\textbf{Data streams.} In order to perform the experimental evaluation of the \textsc{awae}, data streams -- both synthetic and based on real concepts -- with various characteristics were used.

Synthetic data streams were generated using the \emph{Python stream-learn} library\footnote{https://github.com/w4k2/stream-learn}. Three balanced streams were prepared, differing in the concept drift type, and replicated five times based on a random state value to stabilize the results and enable statistical analysis. The following parameters characterized these streams:

\begin{itemize}
    \item \emph{data chunks number} -- 200,
    \item \emph{chunk size} -- 250,
    \item \emph{global label noise} -- $1\%$,
    \item \emph{concept drift type} -- \emph{sudden, gradual, and incremental},
    \item \emph{drifts number} -- 10,
    \item \emph{number of features} -- 8.
\end{itemize}

The second data source was a generator, which enabled the creation of data streams based on static concepts originating from known benchmark static datasets available in data repositories such as \textsc{uci} \cite{Frank:2010} or \textsc{keel} \cite{alcala2011keel}. The datasets used in the stream generation process are presented in Table \ref{tab:DDdatasets}. The generation procedure is based on the temporal interpolation of normalized random projections of original data set into a given subspace, which, depending on the interpolation, may generate sudden drifts (nearest-neighbor interpolation) and incremental drifts (cubic interpolation) \cite{komorniczak2022}.

\begin{table}[!htb]
    \centering
    \vspace{-.5em}
    \caption{Static datasets used for data stream generation.}
    \label{tab:DDdatasets}
    \vspace{-.5em}
\renewcommand{\arraystretch}{1}
\setlength{\tabcolsep}{2pt}

\begin{tabularx}{\columnwidth}{@{}p{4cm}RRR@{}}
\toprule
\bfseries\textsc{dataset} & 
\bfseries\textsc{samples} & 
\bfseries\textsc{features} & 
\bfseries\textsc{classes} \\
\midrule
banknote & 1372 & 4 & 2\\
heart & 270 & 13 & 2 \\
liver & 345 & 6 & 2 \\
monkone & 556 & 6 & 2 \\
sonar & 208 & 60 & 2 \\
soybean & 47 & 35 & 2 \\
wisconsin & 683 & 9 & 2 \\
\bottomrule
\end{tabularx}
\vspace{-1em}
\end{table}

Finally, the experiments were also carried out using 9~real benchmark data streams \cite{Cano:2020,SouzaChallenges:2020} and presented in Table~\ref{tab:real}. Multiclass streams have been binarized and the longest possible fragments have been cut from them, ensuring both classes are present in data chunks containing 250 instances.

\begin{table}[!htb]
    \centering
    \vspace{-.5em}
    \caption{Real data streams characteristics.}
    \label{tab:real}
    \vspace{-.5em}
\renewcommand{\arraystretch}{1}
\setlength{\tabcolsep}{2pt}

\begin{tabularx}{\columnwidth}{@{}p{4cm}RRR@{}}
\toprule
\bfseries\textsc{data stream} & 
\bfseries\textsc{samples} & 
\bfseries\textsc{features} & 
\bfseries\textsc{classes} \\
    \midrule
    INSECTS-abrupt & 48~500 & 33 & 194 \\
    INSECTS-gradual & 21~250 & 33 & 85 \\
    INSECTS-incremental-abrupt & 46~500 & 33 & 186 \\
    INSECTS-incremental-reoccurring & 42~500 & 33 & 170 \\
    INSECTS-incremental & 50~000 & 33 & 200 \\
    airlines & 50~000 & 7 & 200 \\
    covtype & 99~750 & 54 & 399 \\
    electricity & 42~500 & 8 & 170 \\
    poker-lsn & 43~250 & 10 & 173 \\
\bottomrule
\end{tabularx}
\vspace{-1em}
\end{table}

\noindent\textbf{Reference methods.} Throughout the conducted experiments the proposed method was compared with a selection of \emph{state-of-the-art} data stream classification algorithms:
\begin{itemize}
    \item \emph{Streaming Ensemble Algorithm} (\textsc{sea}) \cite{Street2001} -- which trains a new base classifier on each incoming data chunks, and adds it to the maintained classifier pool. In case of exceeding the set maximum pool size, the worst model is removed.
    \item \emph{Accuracy Weighted Ensemble} (\textsc{awe}) \cite{Wang:2003b} -- which gives individual base classifiers a weight on the basis of \emph{mean squared error}.
    \item \emph{Accuracy Updated Ensemble} (\textsc{aue}) \cite{Brzezinski:2011} -- which modifies the \textsc{awe} algorithm  in order to allow updating of base models.
    \item \emph{Learn++ for Non Stationary Environments} (\textsc{nse}) \cite{Ditzler:2013} -- which combines base classifiers using dynamically weighted majority voting, where weights are calculated based on classifiers' errors.
    \item \emph{Online Active Learning Ensemble} (\textsc{oale}) \cite{Shan:2018} -- which employs a long-term stable classifier and multiple dynamic models, supplemented by a hybrid labeling strategy.
\end{itemize}
Each ensemble method contained a base classifier pool with maximum size equal to $10$ model. The following parameters were selected for base classifiers:
\begin{itemize}
    \item \emph{Gaussian Na\"ive Bayes} (\textsc{gnb}) -- where the portion of the largest feature variance that is added to variances for calculation stability is equal to $1e-9$,
    \item \emph{Hoeffding Tree} (\textsc{ht}) -- where the number of instances that should be observed before leaf split attempts is equal to $200$ and \emph{Hellinger distance} is used as a split criterion, 
    \item \emph{Multilayer Perceptron} (\textsc{mlp}) -- with a single hidden layer containing $100$ artificial neurons, \emph{ReLU} activation function, \emph{adam} optimizer, \emph{constant} learning rate, and $200$ maximum iterations.
\end{itemize}

\noindent\textbf{Experimental protocol.} All experiments were conducted using the \emph{stream-learn} library and based on the \emph{Test-Than-Train}~\cite{Krawczyk:2017} evaluation protocol. 

\noindent\textbf{Statistical analysis.} It was performed using the \emph{t-test}~\cite{stapor2021design} during all of the conducted pairwise comparisons. The results of all of the performed tests were reported at a significance level $\alpha = 0.05$.

\noindent \textbf{Reproducibility.} All experiments were conducted using the \emph{stream-learn} library and based on the \emph{Test-Than-Train}~\cite{Krawczyk:2017} evaluation protocol. 
As we mentioned above all ensemble methods have been tested for three types of base classifiers, namely, \emph{Gaussian Naive Bayes} (\textsc{gnb}), \emph{Hoeffding Tree} (\textsc{ht}), and \emph{Multilayer Perceptron} (\textsc{mlp}) according to \emph{scikit-learn} and \emph{scikit-multiflow} implementations \cite{scikit-learn,skmultiflow}. In the case of synthetic streams, \emph{accuracy score} values are reported, while in experiments containing streams based on real concepts -- in order to eliminate the impact of possible data imbalance -- a \emph{balanced accuracy score} was used. The experiments presented in this article can be replicated using the code available in the \emph{GitHub} repository\footnote{\url{https://github.com/w4k2/AWAE}}.

\subsection{Experiment scenarios}
We proposed three groups of experiments to answer the formulated research questions.

\noindent \textbf{Experiment 1 -- Hyperparametrization.} As part of the first experiment, the impact of the values of individual \textsc{awae} parameter pairs on the quality of classification is analyzed in the event of a given type of concept drift on the examples of generated synthetic streams. The result of the experiment is the selection of pseudo-optimal (within the framework of the conducted review) hyperparameter values for \textsc{awae} in scenarios with full label access and with limited labeling.

For all experiments carried out in the restricted access to labels scenario, the following values of the \textsc{bals} algorithm hyperparameters were adopted:
\begin{itemize}
    \item threshold $t$ -- $0.2$,
    \item budget $b$ -- $5\%$.
\end{itemize}
The above values were selected on the basis of previous research results on the \textsc{bals} algorithm.

\noindent \textbf{Experiment 2 -- Comparison with \emph{state-of-the-art} on synthetic streams}

The second experiment presents a comparative analysis of the \textsc{awae} method (in the default and optimized configuration) with \emph{state-of-the-art} methods depending on the type of concept drift and the base classifier used, measuring their performance on the examples of synthetic data streams. The presented results have been extended by statistical analysis using the t-student test. The study covers both scenarios with full and limited labeling.


The \textsc{oale} algorithm in the second experiment was deprived of the \emph{active learning} module to ensure a reliable comparison.

\noindent \textbf{Experiment 3 -- Comparison with \emph{state-of-the-art} on real streams}

The third experiment expands the research from the second one by analyzing real streams with real concepts based on the same pool of comparative methods and identical assumptions. Here, the \textsc{oale} algorithm is used together with described \textsc{bals} approach, which is similar to the active learning technique used originally by the authors of the \emph{Online Active Learning Ensemble}.









\subsection{Experiment 1 -- Finding the best parameter setting}

\emph{Active Weighted Aging Ensemble} is characterized by a reasonably strong hyperparameterization, allowing for the adaptation of the modeling procedure individually to each problem under consideration. Unfortunately, such an approach -- due to evident data peeking coming from fine-tuning -- would hinder the proper comparative assessment of the recognition effectiveness against the \emph{state-of-the-art} methods. Therefore, a preliminary experiment was designed and carried out to select global strategies for scenarios with full and limited labeling. The conducted review indicated, however, that for both of these strategies, the best hyperparameterization is consistent -- showing the same values -- differing in the other terms, like drift type, base classifier, pruning approach and classifier weighting method, therefore this chapter presents only the results for the full labeling scenario.

\begin{figure*}[!htb]
    \centering
    \includegraphics[width=.43\textwidth,clip=true,trim=10 0 15 0]{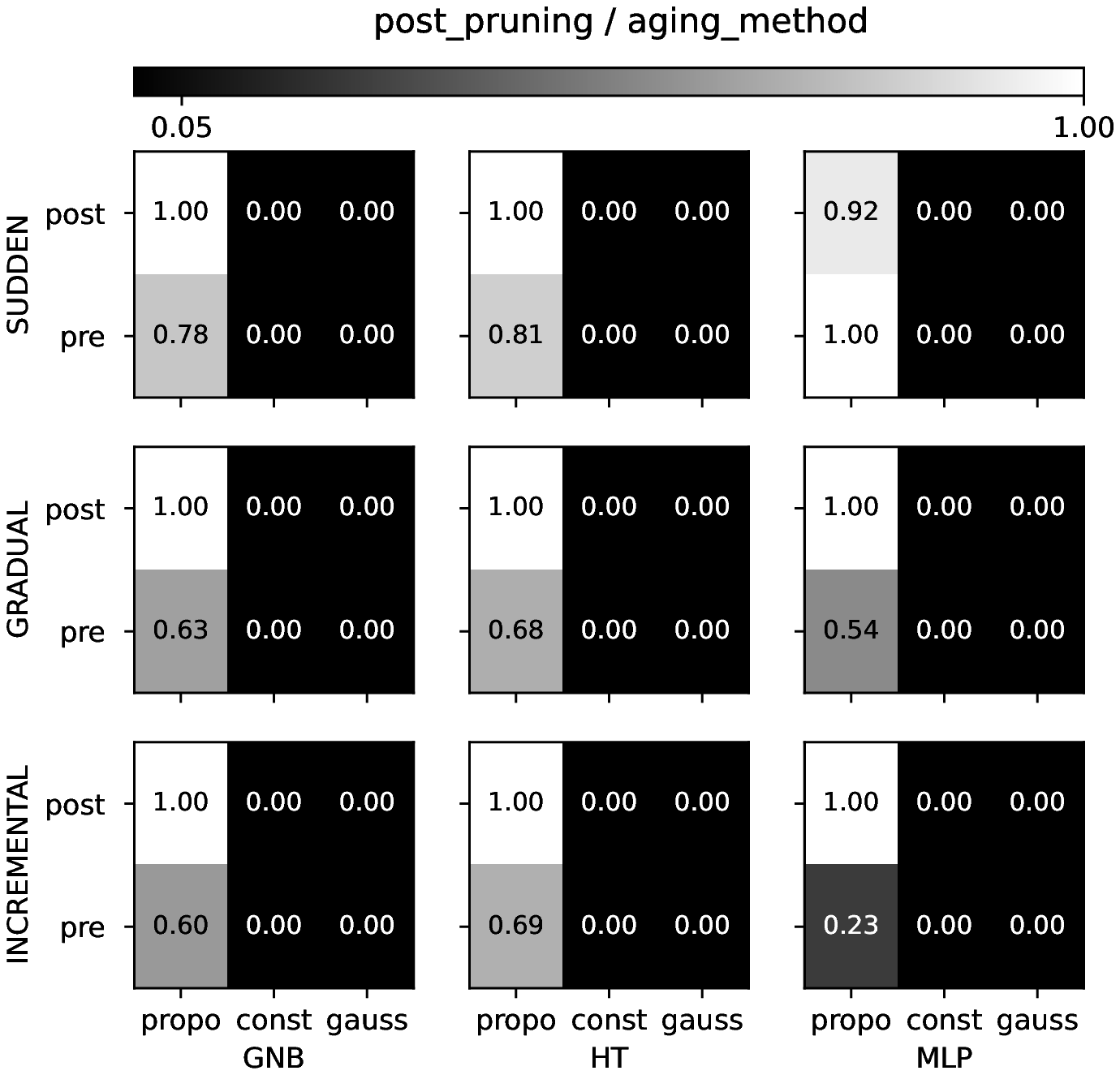}%
    \includegraphics[width=.43\textwidth,clip=true,trim=10 0 15 0]{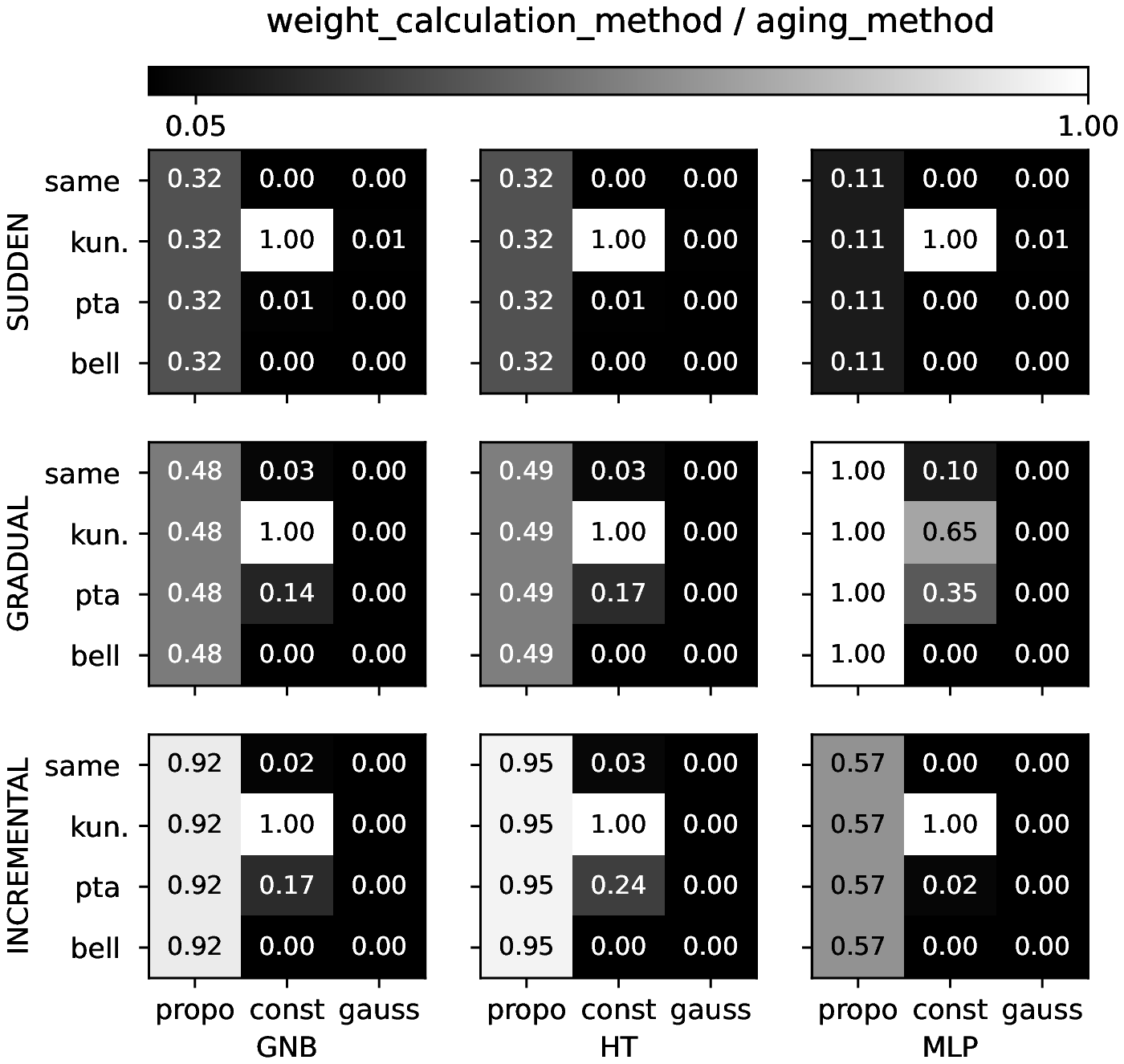}\\
    \includegraphics[width=.43\textwidth,clip=true,trim=10 0 15 0]{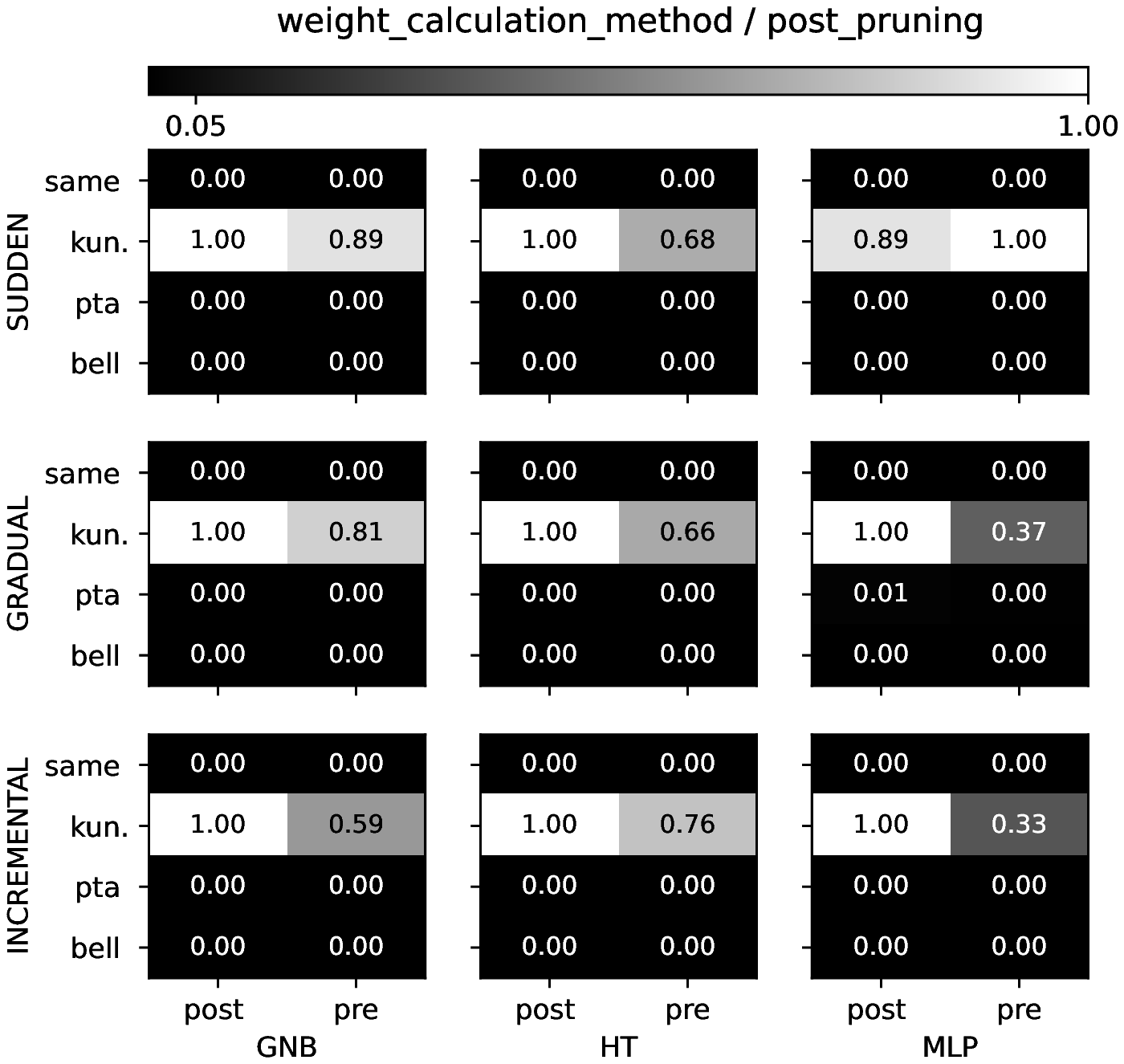}%
    \includegraphics[width=.43\textwidth,clip=true,trim=10 0 15 0]{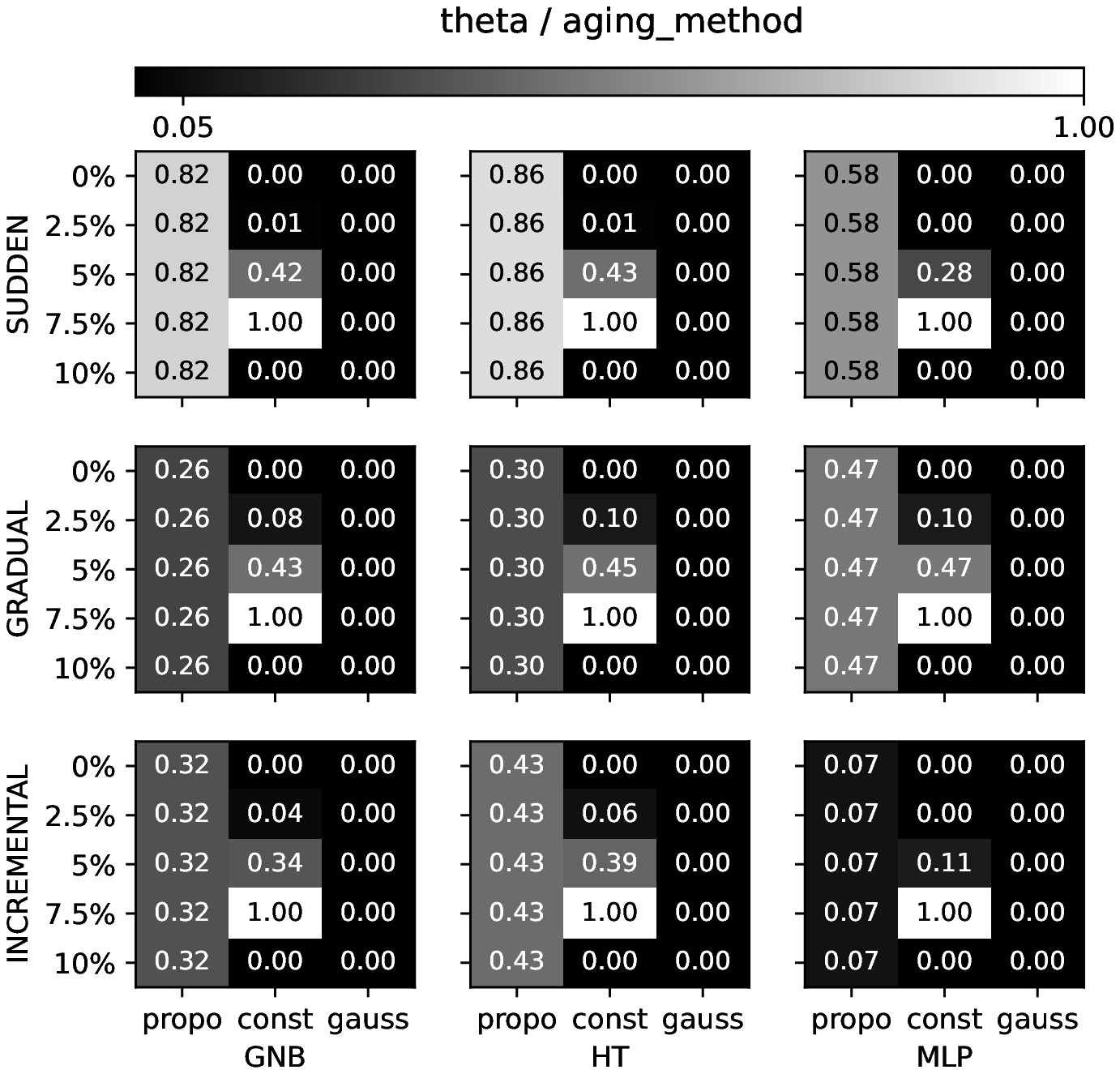}\\
    \includegraphics[width=.43\textwidth,clip=true,trim=10 0 15 0]{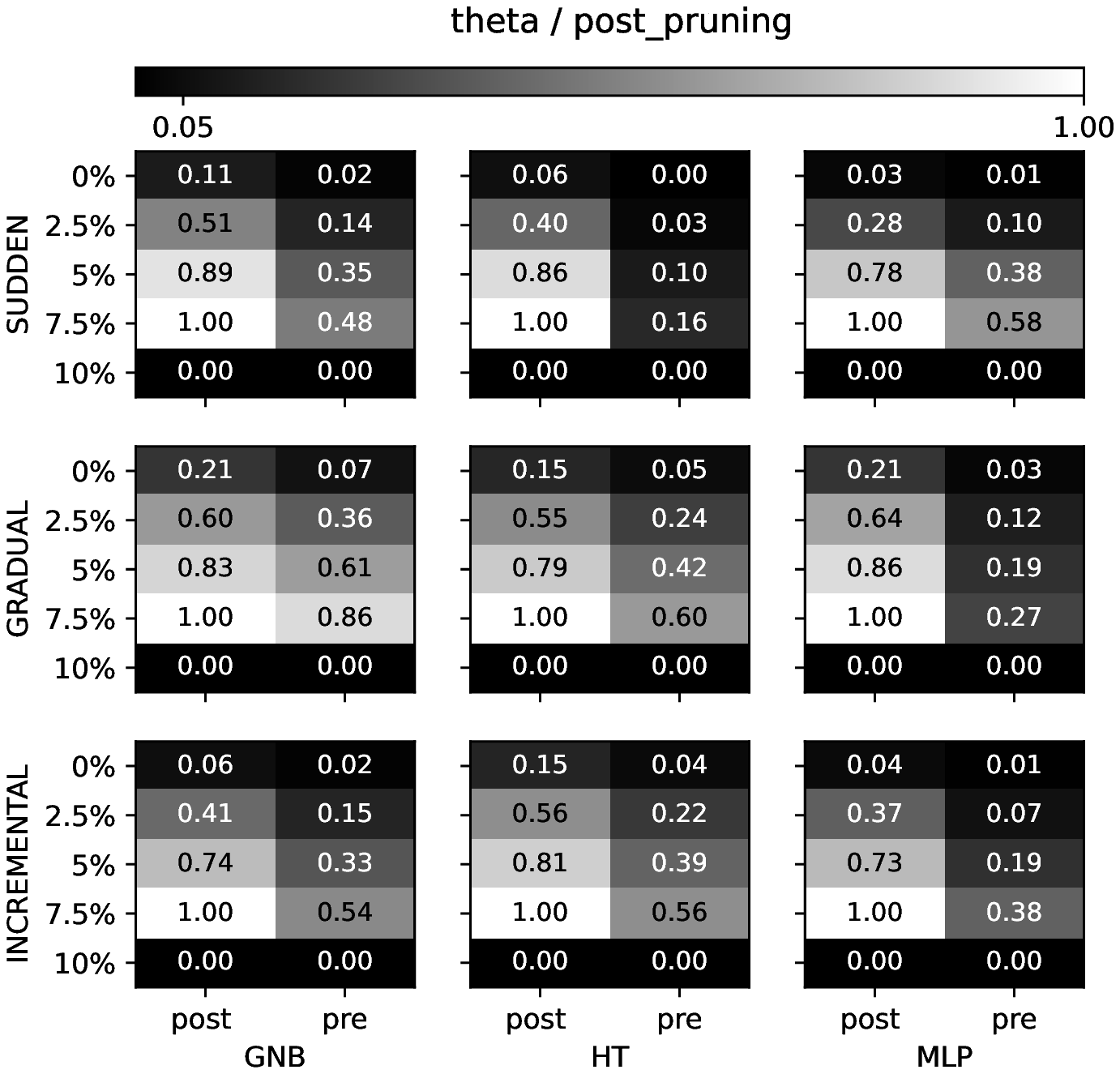}%
    \includegraphics[width=.43\textwidth,clip=true,trim=10 0 15 0]{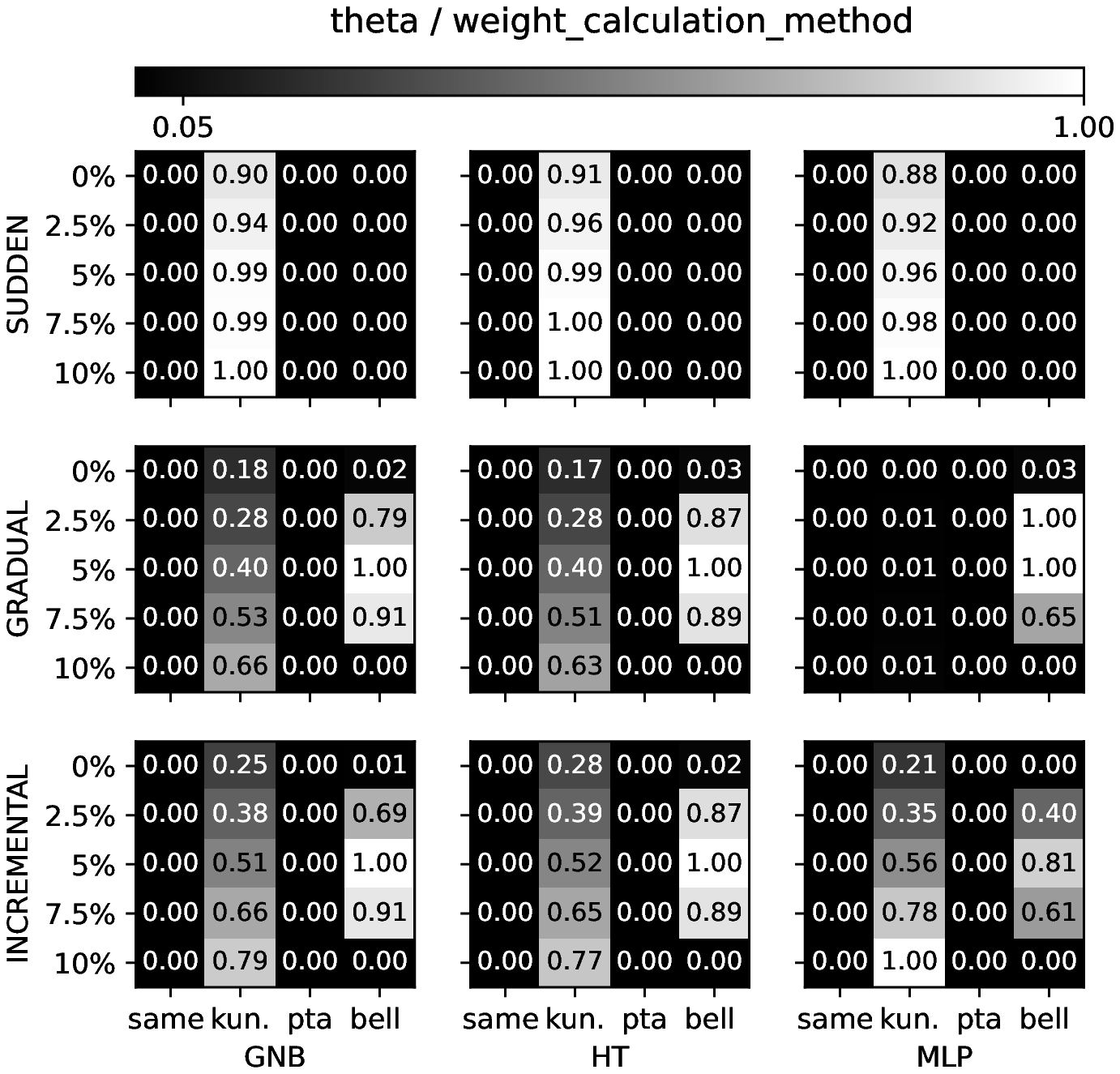}\\
    \caption{\textsc{Experiment 1} --- Graphs of p-values achieved by various configurations of hyperparameter optimization.}
    \label{fig:e1-regular}
\end{figure*}

Even the reduced analysis leads to many tables with the statistical analysis of individual configurations. Therefore the article has been extender with supplementary materials\footnote{\url{https://github.com/w4k2/AWAE/blob/master/supplementary/supplementary.pdf}} in which there is a full set of reference results. This chapter has been limited to the visualization of statistical relationships of models with different configurations (Fig.~\ref{fig:e1-regular}) and table with the selected hyperparameterization for the particular scenarios (Table~\ref{tab:wae_opt_b}). The initial phase of the analysis eliminated the rejuvenation parameter. It combined pruning criterion (Section II.A.3) as essential for processing, leaving four factors tested in six pairs using sudden, gradual and incremental drift using three classification algorithms. The analyzed factors were:

\begin{itemize}
    \item \emph{pruning location} -- post- or pre-pruning (2 values),
    \item \emph{aging} -- proportional, constant and gaussian (3 values),
    \item \emph{weight calculation} -- same, kuncheva, proportional to accuracy and bell curve (4 values),
    \item $\theta$ -- range from 0 to 10\% (5 values).
\end{itemize}

When comparing \emph{pruning location} with the \emph{aging}, a significant advantage of proportional aging was visible, with slight differences between pre- and post-pruning, in some cases showing a statistically insignificant advantage of post-pruning. These relations were not dependent on the drift nature or the base classifier used.

In the juxtaposition of the \emph{weight calculation} with the \emph{aging}, the high stability of the proportional strategy was also evident. Still, it turned out to be slightly inferior, without a significant statistical difference, to the constant strategy paired with the Kuncheva weight calculation method for each case except for gradual drifts using \textsc{mlp} as a base classifier. On the other hand, the comparison of the \emph{weight calculation} with the \emph{pruning location} showed a significant advantage of Kuncheva weight over all other strategies, especially with the application of post-pruning.

Comparing the value of $\theta$ parameter with the \emph{aging} would theoretically suggest the selection of a constant strategy -- leading to the highest results --, but it also shows a strong dependence of this parameter on theta value, building a relatively narrow window in which it is possible to achieve the local optimum. A minimally worse but statistically dependent result, already desensitized to theta value, is achieved with proportional aging, which seems to be a more universal choice of default \textsc{awae} hyperparameterization.

The comparison of the value of $\theta$ parameter with the \emph{pruning location} shows that, up to a certain range, there is a robust and proportional dependence of the model quality on $\theta$ parameter, especially enhanced by post-pruning, which, however, is interrupted suddenly around 0.1 of theta parameter, leading to a significant decrease in predictive ability of a system.

The last comparison between the $\theta$ parameter and the \emph{weight calculation} is the only one that shows the differences in the hyperparameterization of the methods for different drift dynamics. It is clearly visible here that the bell method of weight calculation -- which was not apparent in previous analyzes -- works best for streams with gradual and incremental characteristics, when the Kuncheva method promotes streams with sudden drift.

As can be seen from the above analysis, the available parameterization of the \textsc{awae} method allows for its strong adaptation to the problems. Nevertheless, the quality adjustments introduced by the changes does not always lead to statistically significant changes. Therefore, an attempt was made to select the default configuration, carried out on the review of all 120 combinations, among which the most common among the solutions statistically dependent on the best, for scenarios with full labeling, were the following settings:

\begin{itemize}
    \item $\theta$ -- 5\%,
    \item \emph{weight calculation} -- \emph{bell},
    \item \emph{aging} -- \emph{constant}.
    \item \emph{pruning} -- both pre- and post-pruning were dependent to the best solution.
\end{itemize}

In the case of the active learning scenarios, this hyperparameterization only changed when \emph{Hoeffding Trees} was used as the base classifier. Such configuration is presented in Table~\ref{tab:wae_opt_b}. 

\begin{table}[!ht]
    \centering
    \vspace{-.5em}
    \scriptsize
    \caption{\textsc{Experiment 1} --- \textsc{awae} optimization for active learning.}
    \vspace{-.5em}
\renewcommand{\arraystretch}{1}
\setlength{\tabcolsep}{2pt}

\begin{tabularx}{\columnwidth}{@{}rlCCCCC@{}}
\toprule

& \bfseries Drift type & \bfseries Accuracy & \bfseries Pruning & \bfseries $\theta$ & \bfseries \textsc{wcm} & \bfseries Aging

\\\midrule
\multirow{3}{*}{\rotatebox[origin=c]{90}{\textsc{gnb}}}
&        \textsc{sudden}  &  0.806  &  \textsc{pre}/\textsc{post}  &  5\%  &  \textsc{bell}  &  \textsc{const}  \\
 &       \textsc{gradual}  &  0.788  &  \textsc{pre}/\textsc{post}  &  5\%  &  \textsc{bell}  &  \textsc{const}  \\
&   \textsc{incremental}  &  0.821  &  \textsc{pre}/\textsc{post}  &  5\%  &  \textsc{bell}  &  \textsc{const}  \\\midrule
\multirow{3}{*}{\rotatebox[origin=c]{90}{\textsc{ht}}}
&        \textsc{sudden}  &  0.791  &  \textsc{post}  &  5\%  &  \textsc{pta}  &  \textsc{const}  \\
 &       \textsc{gradual}  &  0.778  &  \textsc{pre}  &  7.5\%  &  \textsc{same}  &  \textsc{const}  \\
&   \textsc{incremental}  &  0.812  &  \textsc{post}  &  7.5\%  &  \textsc{same}  &  \textsc{const}  \\\midrule
\multirow{3}{*}{\rotatebox[origin=c]{90}{\textsc{mlp}}}
 &        \textsc{sudden}  &  0.871  &  \textsc{pre}/\textsc{post}  &  5\%  &  \textsc{bell}  &  \textsc{const}  \\
 &       \textsc{gradual}  &  0.859  &  \textsc{pre}/\textsc{post}  &  5\%  &  \textsc{bell}  &  \textsc{const}  \\
 &   \textsc{incremental}  &  0.887  &  \textsc{pre}/\textsc{post}  &  5\%  &  \textsc{bell}  &  \textsc{const}  \\\bottomrule

    \end{tabularx}
    \vspace{-1em}
    \label{tab:wae_opt_b}
\end{table}

\subsection{Experiment 2 -- Comparison with \emph{state-of-the-art} methods on sythetic streams}

The evaluation of data stream classification methods with the use of synthetic streams, thanks to the possibility of replication of many streams sharing the same characteristics, allows for clear identification of their basic properties, desensitized to the observations of outliers of detailed parameters of the detailed concept. The relevant analysis was carried out in the second experiment for scenarios with full labeling (Table~\ref{tab:ex2_a}) and using active learning (Table~\ref{tab:ex2_b}).

\begin{table}[!htb]
    \centering
    \scriptsize
    \caption{\textsc{Experiment 2} --- final results on synthetic problems for regular scenarios. Small numbers indicates the indexes of methods from which the examined one is statistically better.}
    \renewcommand{\arraystretch}{1}
\setlength{\tabcolsep}{2pt}

\begin{tabularx}{\textwidth}{@{}cCCC@{\hspace{1em}}|@{\hspace{1em}}CCC@{\hspace{1em}}|@{\hspace{1em}}CCC@{}}
\toprule

& \multicolumn{3}{c@{\hspace{1em}}|@{\hspace{1em}}}{\bfseries Sudden drift} & \multicolumn{3}{c@{\hspace{1em}}|@{\hspace{1em}}}{\bfseries Gradual drift} & \multicolumn{3}{c}{\bfseries Incremental drift}  \\
& \bfseries\textsc{gnb} & \bfseries\textsc{ht} & \bfseries\textsc{mlp} & \bfseries\textsc{gnb} & \bfseries\textsc{ht} & \bfseries\textsc{mlp}& \bfseries\textsc{gnb} & \bfseries\textsc{ht} & \bfseries\textsc{mlp}\\\midrule

\textsc{sea}

&0.801 & 0.802 & 0.859 & 0.813 & 0.813 & 0.904 & 0.843 & 0.843 & 0.896 \\
$^{(1)}$ & $^{-}$ & $^{-}$ & $^{5}$ & $^{-}$ & $^{-}$ & $^{5}$ & $^{-}$ & $^{-}$ & $^{5}$ \\

\textsc{awe}
& 0.802 & 0.803 & 0.860 & 0.813 & 0.813 & 0.904 & 0.843 & 0.840 & 0.896 \\
$^{(2)}$ & $^{-}$ & $^{-}$ & $^{5}$ & $^{-}$ & $^{-}$ & $^{5}$ & $^{-}$ & $^{-}$ & $^{5}$ \\

\textsc{aue}
& 0.832 & 0.872 & 0.874 & 0.833 & 0.858 & 0.909 & 0.865 & 0.892 & 0.904 \\
$^{(3)}$ & $^{1, 2}$ & $^{all}$ & $^{1, 2, 5}$ & $^{1, 2, 5}$ & $^{1, 2, 4, 6}$ & $^{5}$ & $^{1, 2, 5}$ & $^{all}$ & $^{5}$ \\

\textsc{nse}
& 0.860 & 0.860 & 0.922 & 0.841 & 0.841 & 0.920 & 0.872 & 0.871 & 0.929 \\
$^{(4)}$ & $^{1, 2, 3, 5}$ & $^{1, 2}$ & $^{1, 2, 3, 5}$ & $^{1, 2, 5}$ & $^{1, 2}$ & $^{1, 2, 3, 5}$ & $^{1, 2, 5}$ & $^{1, 2}$ & $^{1, 2, 3, 5}$ \\

\textsc{oale}
& 0.827 & 0.861 & 0.703 & 0.817 & 0.854 & 0.722 & 0.850 & 0.874 & 0.712 \\
$^{(5)}$ & $^{1, 2}$ & $^{1, 2}$ & $^{-}$ & $^{-}$ & $^{1, 2, 4, 6}$ & $^{-}$ & $^{-}$ & $^{1, 2}$ & $^{-}$ \\

\textsc{awae} & 0.858 & 0.858 & 0.934 & 0.843 & 0.841 & 0.928 & 0.876 & 0.873 & 0.944 \\
$^{(6)}$ & $^{1, 2, 3, 5}$ & $^{1, 2}$ & $^{all}$ & $^{1, 2, 3, 5}$ & $^{1, 2}$ & $^{all}$ & $^{1, 2, 3, 5}$ & $^{1, 2}$ & $^{all}$ \\

\bottomrule

\end{tabularx}

\vspace{-1em}
    \label{tab:ex2_a}
\end{table}

In the case of the full labeled synthetic streams, \textsc{mlp} clearly turns out to be the best classification algorithm, which allows for the highest classification quality for each type of drift. Additionally, it is the algorithm that works best when paired with the proposed \textsc{awae} algorithm. Both in the case of sudden, gradual, and incremental drifts, it achieves a significant statistical advantage over any of the \emph{state-of-the-art} methods. Relatively the worst is the match between \textsc{ht-awae}, where the statistical advantage is achieved only over the \textsc{sea}, and \textsc{awe} methods, and the \textsc{ht-aue} match is the best, which is still statistically significantly worse than the \textsc{mlp-awae}. The \textsc{gnb-awae} pair allows obtaining results dependent on \textsc{nse}, with which together it is the best choice when using the \emph{Gaussian Naive Bayes} classifier.

\begin{table}[!htb]
    \centering
    \scriptsize
    \caption{\textsc{Experiment 2} --- final results on synthetic problems for active learning scenarios. Small numbers indicates the indexes of methods from which the examined one is statistically better.}
    \renewcommand{\arraystretch}{1}
\setlength{\tabcolsep}{2pt}

\begin{tabularx}{\textwidth}{@{}cCCC@{\hspace{1em}}|@{\hspace{1em}}CCC@{\hspace{1em}}|@{\hspace{1em}}CCC@{}}
\toprule

& \multicolumn{3}{c@{\hspace{1em}}|@{\hspace{1em}}}{\bfseries Sudden drift} & \multicolumn{3}{c@{\hspace{1em}}|@{\hspace{1em}}}{\bfseries Gradual drift} & \multicolumn{3}{c}{\bfseries Incremental drift}  \\
& \bfseries\textsc{gnb} & \bfseries\textsc{ht} & \bfseries\textsc{mlp} & \bfseries\textsc{gnb} & \bfseries\textsc{ht} & \bfseries\textsc{mlp}& \bfseries\textsc{gnb} & \bfseries\textsc{ht} & \bfseries\textsc{mlp}\\\midrule

\textsc{sea}
&0.842 & 0.843 & 0.871 & 0.856 & 0.855 & 0.909 & 0.886 & 0.885 & 0.920 \\
$^{(1)}$ & $^{4, 5}$ & $^{4, 5}$ & $^{4, 5}$ & $^{4, 5}$ & $^{4, 5}$ & $^{4, 5}$ & $^{4, 5}$ & $^{4, 5}$ & $^{4, 5}$ \\

\textsc{awe}
& 0.841 & 0.842 & 0.870 & 0.855 & 0.854 & 0.908 & 0.885 & 0.886 & 0.919 \\
$^{(2)}$ & $^{4, 5}$ & $^{4, 5}$ & $^{4, 5}$ & $^{4, 5}$ & $^{4, 5}$ & $^{4, 5}$ & $^{4, 5}$ & $^{4, 5}$ & $^{4, 5}$ \\

\textsc{aue}
& 0.874 & 0.884 & 0.871 & 0.863 & 0.873 & 0.909 & 0.898 & 0.903 & 0.920 \\
$^{(3)}$ & $^{1, 2, 4, 5}$ & $^{all}$ & $^{4, 5}$ & $^{4, 5}$ & $^{1, 2, 4, 5}$ & $^{4, 5}$ & $^{1, 2, 4, 5}$ & $^{1, 2, 4, 5}$ & $^{4, 5}$ \\

\textsc{nse}
& 0.536 & 0.540 & 0.586 & 0.540 & 0.550 & 0.595 & 0.539 & 0.541 & 0.598 \\
$^{(4)}$ & $^{-}$ & $^{-}$ & $^{-}$ & $^{-}$ & $^{-}$ & $^{-}$ & $^{-}$ & $^{-}$ & $^{-}$ \\

\textsc{oale}
& 0.788 & 0.789 & 0.604 & 0.773 & 0.780 & 0.608 & 0.812 & 0.814 & 0.608 \\
$^{(5)}$ & $^{4}$ & $^{4}$ & $^{-}$ & $^{4}$ & $^{4}$ & $^{-}$ & $^{4}$ & $^{4}$ & $^{-}$  \\

\textsc{awae}
& 0.866 & 0.868 & 0.912 & 0.866 & 0.866 & 0.923 & 0.896 & 0.896 & 0.941 \\
$^{(6)}$ & $^{1, 2, 4, 5}$ & $^{1, 2, 4, 5}$ & $^{all}$ & $^{4, 5}$ & $^{4, 5}$ & $^{all}$ & $^{2, 4, 5}$ & $^{4, 5}$ & $^{all}$ \\


\bottomrule
\end{tabularx}

    \label{tab:ex2_b}
\end{table}

The results for the active learning strategy are similar. Here also globally the best combination is \textsc{mlp-awae}.

\subsection{Experiment 3 -- Comparison with state-of-the-art on real streams}

\begin{figure*}
    \centering
    \vspace{1em}
    \includegraphics[width=.9\textwidth,clip=true,trim=0 505 0 0]{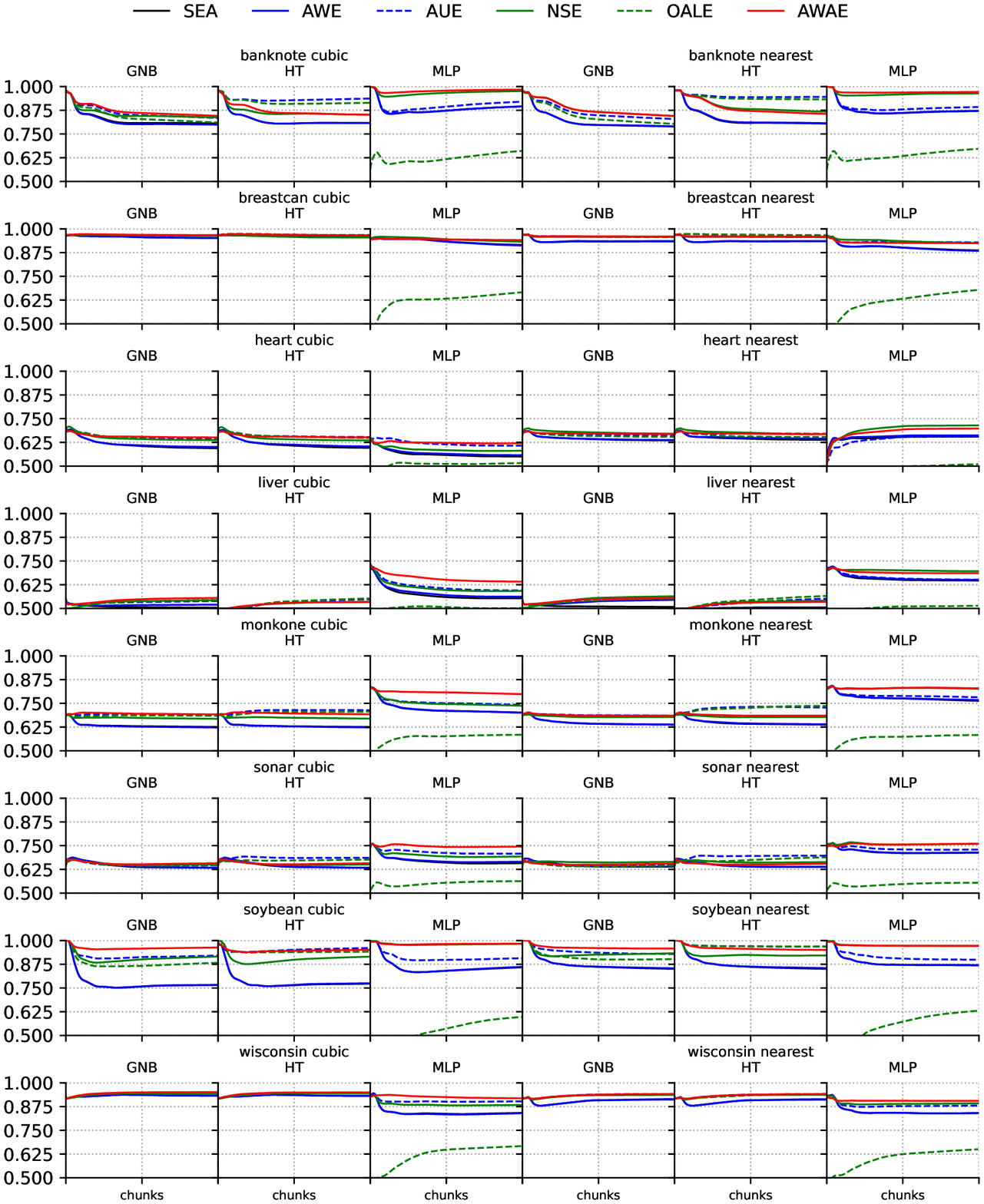}
    \vspace{2em}\;
    \includegraphics[width=.9\textwidth,clip=true,trim=0 505 0 35]{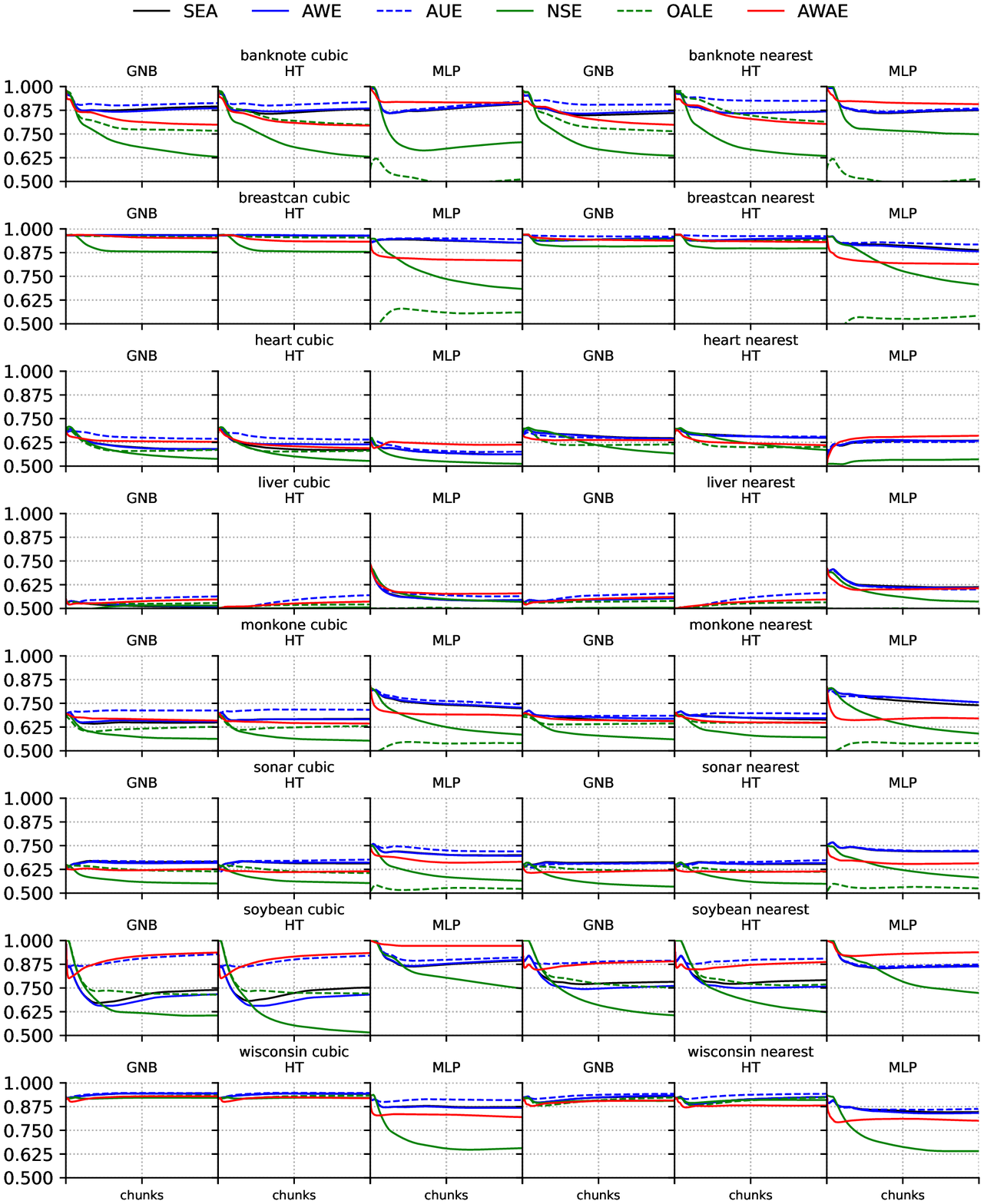}
    \caption{\textsc{Experiment 3} --- final results on selected streams with real concepts for (top two) regular and (bottom two) active learning scenarios.}
    \label{fig:my_label_a}
\end{figure*}

\begin{figure*}[!htb]
    \centering
    \includegraphics[width=.49\textwidth,clip=true,trim=0 380 0 0]{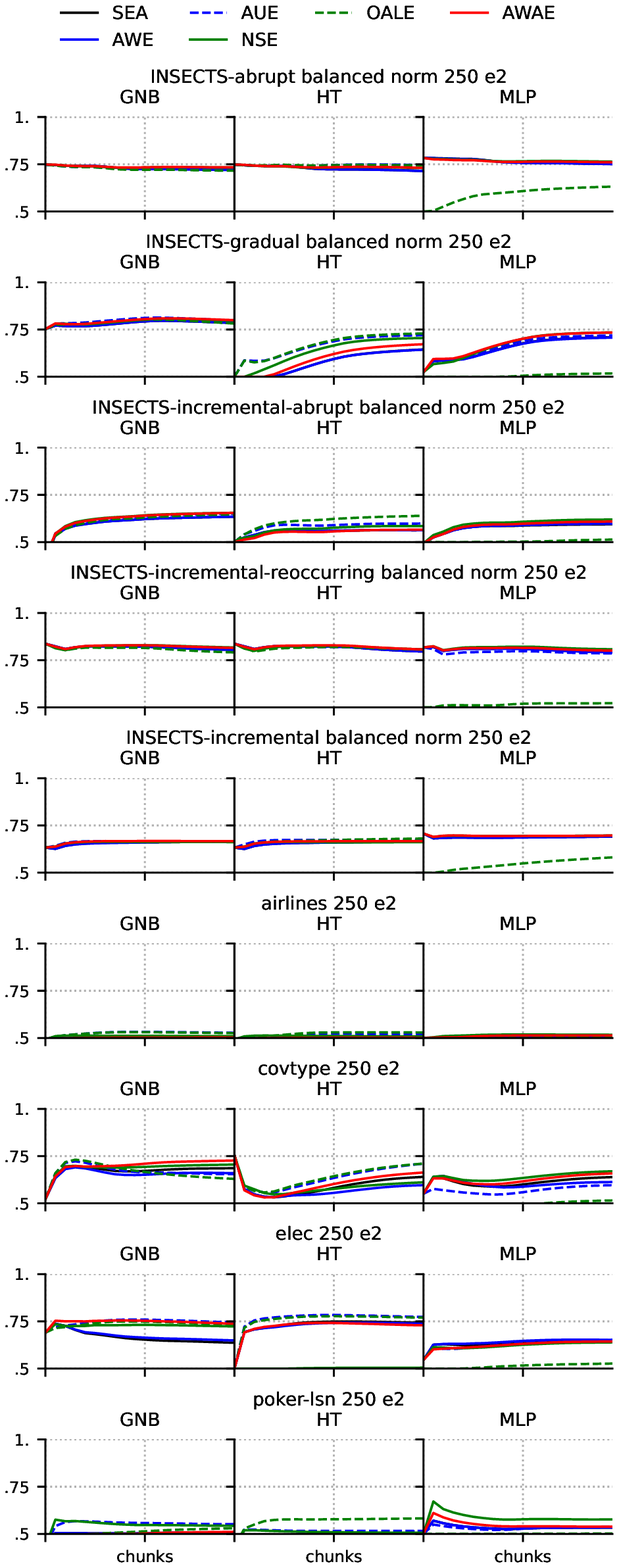}%
    \includegraphics[width=.49\textwidth,clip=true,trim=0 380 0 0]{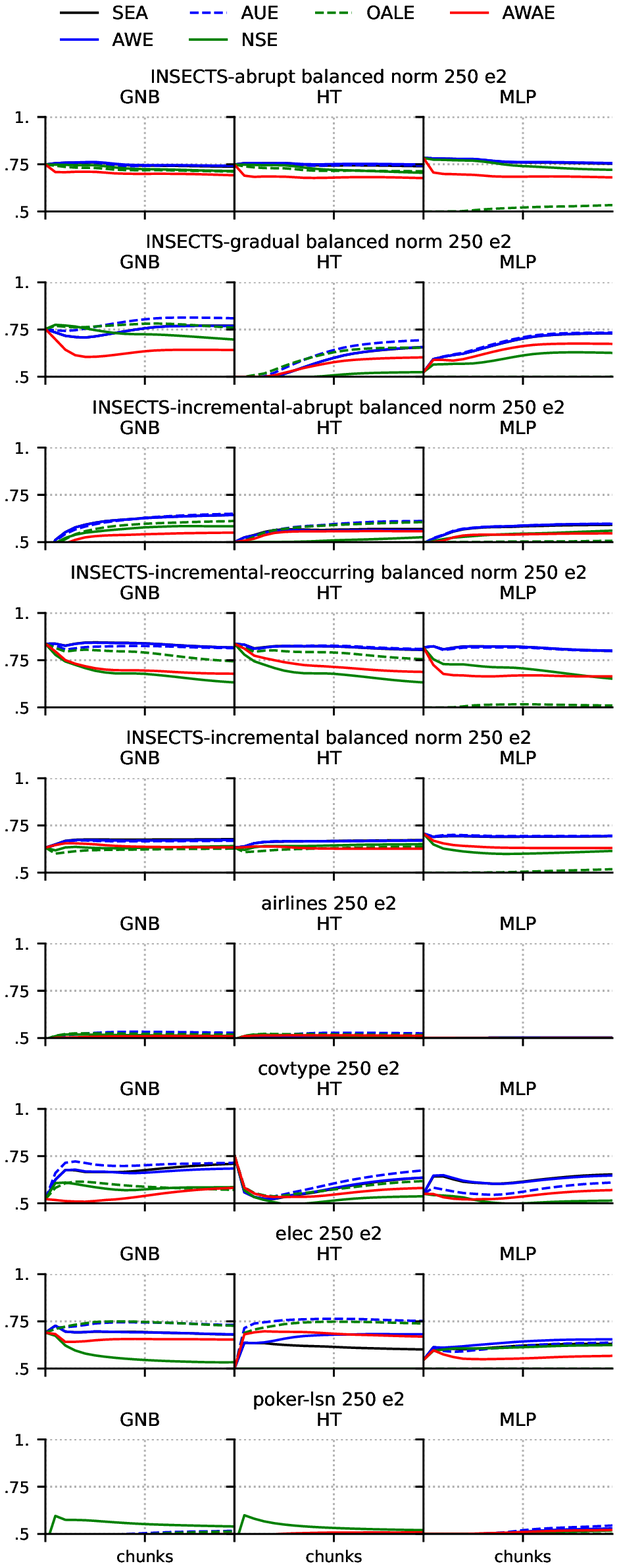}
    \caption{\textsc{Experiment 3} --- final results on selected real streams for regular (left) and active learning scenarios (right).}
    \label{fig:my_label_b}
\end{figure*}

The presentation of the processing efficiency of classification algorithms in the case of real streams is optimal when visualizing the accumulative sums of the flow efficiency, thanks to which both the dynamics of learning new concepts and the possible tendency to reduce the discriminative abilities of the recognition system in the course of neutralization to successive drifts are clearly visible. Therefore, the experimental evaluation of the \textsc{awae} algorithm on streams with real concepts (Figure~\ref{fig:my_label_a}) and real streams (Figure~\ref{fig:my_label_b}) was carried out using appropriate plots.

In the case of the classification of streams with real concepts, the observations from Experiment 2 are replicated, where the \textsc{mlp-awae} combination turned out to be the best match. In the case of learning with full labeling, \textsc{mlp-nse} is highly competitive with it, and \textsc{ht-aue} is equal competition in the case of active learning scenarios. This is visible in both cubic and nearest drift.

The only cases in which the \textsc{mlp-awae} combination do not turn out to be the optimal or optimal-dependent processing strategy is a part of real streams, in particular from the \textsc{insects} group, where it differs significantly from other comparative methods. It probably results from the different characteristics of the concept, where the default parameterization defined by the outcome of Experiment~1 turns out to be a wrong strategy. 

\subsection{Lessons learned}

It can be seen that \textsc{awae} establishes similar parameters when dealing with the full labeled data stream. The preference is constant aging with $\theta=5\%$ and the method of weight calculating consistent with Eq. 6. For the active learning scenario, we may observe that similar parameter values are determined when \textsc{mlp} and \textsc{gnb} are chosen as base classifiers. The situation is different when \textsc{ht} is selected when the suggested parameters strongly depend on the drift type present (RQ1 answered).
\textsc{awae} can work with any type of base classifier, but in the experiments, it was limited to only three methods, of which the best quality is obtained by combining \textsc{awae} with \textsc{mlp} (RQ2 answered).

The same combination also performs best when compared to \emph{state-of-the-art} methods. For synthetic streams, for all drift types, we observe statistically significantly better quality of \textsc{mlp-awae} classification compared to reference methods using \textsc{mlp} as base classifier. In the case of using \textsc{gnb}, \textsc{awae} is statistically significantly better than all methods except \textsc{nse}, which also for this classifier achieves better quality than the other methods. When \textsc{ht} is used, \textsc{aue} is by far the best method, especially in the case of sudden drift. However, it should be noted that a similar relationship can be observed for real data streams (RQ3 and RQ4 answered). Similar observations can be made for real data streams when the active learning strategy \textsc{bals} is applied, the \textsc{mlp-awae} performs best. The exceptions are selected streams from the \textsc{insects} group, where \textsc{awae} performs quite average, while the best quality is mainly characterized by \textsc{aue} and \textsc{awe}. For all real data streams, \textsc{olae} and \textsc{nse} perform the worst for scenarios with \textsc{bals}. However, it should be noted here that \textsc{olae} has its own built-in active learning method that \textsc{bals} replaced for the experiments, which did not perform well for it and making a general conclusion about \textsc{olae} quality in the case of its modification would be unfair (RQ5 answered).

To summarize, the \textsc{awae} algorithm can be recommended as an effective tool for data stream processing, emphasizing, however, that this method is quite sensitive to configuration and requires special attention when used for a particular task. It is visible for some real data streams, where a lack of appropriate parameter settings may deteriorate the predictive performance.

\section{Conclusion}

This study aimed to develop a novel, effective framework for a drifted data stream classification task. We proposed the \emph{Active Weighted Aging Ensemble} algorithm that utilizes the changing ensemble lineup to appropriately react to \emph{concept drift} and active learning strategy to reduce the budget required for upcoming instance labeling.
The research conducted on benchmark data streams confirmed the effectiveness of the proposed solution. It highlighted its strengths in comparison with \emph{state-of-the-art} methods. It is also worth mentioning that estimated computational complexity is acceptable and comparable to the benchmark algorithms.
This work is a step forward towards using active and ensemble learning to design effective classification models for drifted data streams. The obtained results encourage us to continue working on this concept. Future research may include:

\begin{itemize}
    \item Application of the proposed method to imbalanced data stream classification, especially considering the application of data preprocessing for data balancing.
    \item Employing a drift detection techniques to speed-up the reaction to the \emph{concept drift}, especially in the case of abrupt changes.
    \item Evaluation of how \textsc{awae} is robust to different distributions of the label and feature noise.
    \item Using \textsc{awae} on massive and high-speed data streams requires a deeper study on the effective ways of its parallelization.
    \item Proposing an effective active learning strategy for multi-modal data stream processing, especially for possible label propagation among modalities.
    \item Application of \textsc{awae} to a real-world data stream susceptible to the presence of data stream, i.e., medical or banking data.
\end{itemize}

\section*{Acknowledgement}
This work is supported by the CEUS-UNISONO programme, which has received funding from the National Science Centre, Poland under grant agreement No. 2020/02/Y/ST6/00037.

%
%
%
\bibliographystyle{splncs04}
\bibliography{bibliography}

\begin{thebibliography}{10}
\providecommand{\url}[1]{\texttt{#1}}
\providecommand{\urlprefix}{URL }
\providecommand{\doi}[1]{https://doi.org/#1}

\bibitem{alcala2011keel}
Alcal{\'a}-Fdez, J., Fern{\'a}ndez, A., Luengo, J., Derrac, J., Garc{\'\i}a,
  S., S{\'a}nchez, L., Herrera, F.: Keel data-mining software tool: data set
  repository, integration of algorithms and experimental analysis framework.
  Journal of Multiple-Valued Logic \& Soft Computing  \textbf{17} (2011)

\bibitem{Attenberg:2011}
Attenberg, J., Provost, F.: Online active inference and learning. In:
  Proceedings of the 17th ACM SIGKDD International Conference on Knowledge
  Discovery and Data Mining. p. 186–194. KDD '11, Association for Computing
  Machinery, New York, NY, USA (2011). \doi{10.1145/2020408.2020443},
  \url{https://doi.org/10.1145/2020408.2020443}

\bibitem{baena2006early}
Baena-Garc{\i}a, M., del Campo-{\'A}vila, J., Fidalgo, R., Bifet, A., Gavalda,
  R., Morales-Bueno, R.: Early drift detection method. In: Fourth international
  workshop on knowledge discovery from data streams. vol.~6, pp. 77--86 (2006)

\bibitem{Baidari:2020}
Baidari, I., Honnikoll, N.: Accuracy weighted diversity-based online boosting.
  Expert Systems with Applications  \textbf{160},  113723 (07 2020).
  \doi{10.1016/j.eswa.2020.113723}

\bibitem{Barros:2016}
Barros, R.S.M.d., Garrido T.~de Carvalho~Santos, S., Gonçalves~Júnior, P.M.:
  A boosting-like online learning ensemble. In: 2016 International Joint
  Conference on Neural Networks (IJCNN). pp. 1871--1878 (2016).
  \doi{10.1109/IJCNN.2016.7727427}

\bibitem{Bifet:2006}
Bifet, A., Gavald{\`a}, R.: Kalman filters and adaptive windows for learning in
  data streams. In: Todorovski, L., Lavra{\v{c}}, N., Jantke, K.P. (eds.)
  Discovery Science. pp. 29--40. Springer Berlin Heidelberg, Berlin, Heidelberg
  (2006)

\bibitem{BifetGavalda:2007}
Bifet, A., Gavald{\`{a}}, R.: Learning from time-changing data with adaptive
  windowing. In: Proceedings of the Seventh {SIAM} International Conference on
  Data Mining, April 26-28, 2007, Minneapolis, Minnesota, {USA}. pp. 443--448.
  {SIAM} (2007). \doi{10.1137/1.9781611972771.42},
  \url{https://doi.org/10.1137/1.9781611972771.42}

\bibitem{Bifet:2007}
Bifet, A., Gavalda`, R.: Learning from time-changing data with adaptive
  windowing. In: Proceedings of the Seventh SIAM International Conference on
  Data Mining, April 26-28, 2007, Minneapolis, Minnesota, USA. SIAM (2007)

\bibitem{Bifet:2010LevBag}
Bifet, A., Holmes, G., Pfahringer, B.: Leveraging bagging for evolving data
  streams. In: ECML/PKDD (1). pp. 135--150 (2010)

\bibitem{Blanco:2015}
Blanco, I.I.F., del Campo{-}Avila, J., Ramos{-}Jimenez, G., Bueno, R.M., Diaz,
  A.A.O., Mota, Y.C.: Online and non-parametric drift detection methods based
  on hoeffding's bounds. {IEEE} Trans. Knowl. Data Eng.  \textbf{27}(3),
  810--823 (2015)

\bibitem{Bouguelia:2016}
Bouguelia, M.R., Bela\"{\i}d, Y., Bela\"{\i}d, A.: An adaptive streaming active
  learning strategy based on instance weighting. Pattern Recogn. Lett.
  \textbf{70}(C),  38–44 (jan 2016). \doi{10.1016/j.patrec.2015.11.010},
  \url{https://doi.org/10.1016/j.patrec.2015.11.010}

\bibitem{Brzezinski:2011}
Brzezi{\'{n}}ski, D., Stefanowski, J.: Accuracy updated ensemble for data
  streams with concept drift. In: Corchado, E., Kurzy{\'{n}}ski, M.,
  Wo{\'{z}}niak, M. (eds.) Hybrid Artificial Intelligent Systems. pp. 155--163.
  Springer Berlin Heidelberg, Berlin, Heidelberg (2011)

\bibitem{Cano:2020}
Cano, A., Krawczyk, B.: Kappa updated ensemble for drifting data stream mining.
  Mach. Learn.  \textbf{109}(1),  175--218 (2020).
  \doi{10.1007/s10994-019-05840-z},
  \url{https://doi.org/10.1007/s10994-019-05840-z}

\bibitem{Cohen:2003}
Cohen, E., Strauss, M.: Maintaining time-decaying stream aggregates. In:
  Proceedings of the Twenty-Second ACM SIGMOD-SIGACT-SIGART Symposium on
  Principles of Database Systems. p. 223–233. PODS '03, Association for
  Computing Machinery, New York, NY, USA (2003). \doi{10.1145/773153.773175},
  \url{https://doi.org/10.1145/773153.773175}

\bibitem{Ditzler:2013}
Ditzler, G., Polikar, R.: Incremental learning of concept drift from streaming
  imbalanced data. IEEE Transactions on Knowledge and Data Engineering
  \textbf{25}(10),  2283--2301 (Oct 2013)

\bibitem{Domingos:2003}
Domingos, P., Hulten, G.: A general framework for mining massive data streams.
  Journal of Computational and Graphical Statistics  \textbf{12},  945--949
  (2003)

\bibitem{du2014selective}
Du, L., Song, Q., Zhu, L., Zhu, X.: A selective detector ensemble for concept
  drift detection. The Computer Journal p. bxu050 (2014)

\bibitem{Duda:2001}
Duda, R.O., Hart, P.E., Stork, D.G.: Pattern Classification. Wiley, New York,
  2. edn. (2001)

\bibitem{Frank:2010}
Frank, A., Asuncion, A.: {UCI} machine learning repository,
  http://archive.ics.uci.edu/ml (2010)

\bibitem{Gama:2013}
Gama, J., Zliobaite, I., Bifet, A., Pechenizkiy, M., Bouchachia, A.: A survey
  on concept drift adaptation. ACM Computing Surveys  \textbf{in press} (2013)

\bibitem{GamaUFFT:2004}
Gama, J.a., Medas, P., Rocha, R.: Forest trees for on-line data. In:
  Proceedings of the 2004 ACM Symposium on Applied Computing. p. 632–636. SAC
  '04, Association for Computing Machinery, New York, NY, USA (2004).
  \doi{10.1145/967900.968033}, \url{https://doi.org/10.1145/967900.968033}

\bibitem{Gama:2004}
Gama, J., Medas, P., Castillo, G., Rodrigues, P.: Learning with drift
  detection. In: In SBIA Brazilian Symposium on Artificial Intelligence. pp.
  286--295. Springer Verlag (2004)

\bibitem{Greiner:2002}
Greiner, R., Grove, A.J., Roth, D.: Learning cost-sensitive active classifiers.
  Artif. Intell.  \textbf{139}(2),  137--174 (Aug 2002)

\bibitem{Gustafsson:2000}
Gustafsson, F.: {Adaptive Filtering and Change Detection}. Wiley (Oct 2000)

\bibitem{Hoens:2012}
Hoens, T.R., Polikar, R., Chawla, N.V.: Learning from streaming data with
  concept drift and imbalance: an overview. Prog. Artif. Intell.
  \textbf{1}(1),  89--101 (2012). \doi{10.1007/s13748-011-0008-0},
  \url{https://doi.org/10.1007/s13748-011-0008-0}

\bibitem{Huang:2013}
Huang, D., Koh, Y.S., Dobbie, G., Pears, R.: Tracking drift types in changing
  data streams. vol.~8346, pp. 72--83 (12 2013)

\bibitem{Jacobs:1991}
Jacobs, R.A., Jordan, M.I., Nowlan, S.J., Hinton, G.E.: Adaptive mixtures of
  local experts. Neural Comput.  \textbf{3},  79--87 (March 1991)

\bibitem{Junsawang:2019}
Junsawang, P., Phimoltares, S., Lursinsap, C.: Streaming chunk incremental
  learning for class-wise data stream classification with fast learning speed
  and low structural complexity. PLOS ONE  \textbf{14}(9),  1--20 (09 2019).
  \doi{10.1371/journal.pone.0220624},
  \url{https://doi.org/10.1371/journal.pone.0220624}

\bibitem{Kolter:2003}
Kolter, J., Maloof, M.: Dynamic weighted majority: a new ensemble method for
  tracking concept drift. In: Data Mining, 2003. ICDM 2003. Third IEEE
  International Conference on. pp. 123 -- 130 (nov 2003)

\bibitem{komorniczak2022}
Komorniczak, J., Ksieniewicz, P.: Data stream generation through real concept's
  interpolation. In: unpublished

\bibitem{Korycki:2018}
Korycki, {\L}., Krawczyk, B.: Combining Active Learning and Self-Labeling for
  Data Stream Mining, pp. 481--490. Springer International Publishing, Cham
  (2018)

\bibitem{Krawczyk:2017}
Krawczyk, B., Minku, L.L., Gama, J., Stefanowski, J., Woźniak, M.: Ensemble
  learning for data stream analysis: A survey. Information Fusion  \textbf{37},
   132--156 (2017). \doi{https://doi.org/10.1016/j.inffus.2017.02.004},
  \url{https://www.sciencedirect.com/science/article/pii/S1566253516302329}

\bibitem{Ksieniewicz:2019}
Ksieniewicz, P., Woźniak, M., Cyganek, B., Kasprzak, A., Walkowiak, K.: Data
  stream classification using active learned neural networks. Neurocomputing
  \textbf{353},  74--82 (2019).
  \doi{https://doi.org/10.1016/j.neucom.2018.05.130},
  \url{https://www.sciencedirect.com/science/article/pii/S0925231219303248},
  recent Advancements in Hybrid Artificial Intelligence Systems

\bibitem{Kuncheva:2004MCS}
Kuncheva, L.I.: Classifier ensembles for changing environments. In: Proc. 5th
  MCS Int. Workshop on Mult. Class. Syst. Lecture Notes in Computer Science,
  vol.~3077, pp. 1--15. Springer (2004)

\bibitem{Kuncheva:2014}
Kuncheva, L.I.: Combining Pattern Classifiers: Methods and Algorithms - second
  edition. Wiley-Interscience (2014)

\bibitem{Kurlej:2011}
Kurlej, B., Wozniak, M.: Impact of window size in active learning of evolving
  data streams. In: Proceedings of the 45th International Conference on
  Modelling and Simulation of Systems MOSIS 2011. pp. 56--62 (2011)

\bibitem{Kurlej:2012}
Kurlej, B., Wozniak, M.: Active learning approach to concept drift problem.
  Logic Journal of the IGPL  \textbf{20}(3),  550--559 (2012)

\bibitem{Lan:2009}
Lan, Y., Soh, Y.C., Huang, G.: Ensemble of online sequential extreme learning
  machine. Neurocomputing  \textbf{72}(13-15),  3391--3395 (2009)

\bibitem{Lazarescu:2004}
Lazarescu, M.M., Venkatesh, S., Bui, H.H.: Using multiple windows to track
  concept drift. Intell. Data Anal.  \textbf{8}(1),  29--59 (Jan 2004)

\bibitem{Lee:2004}
Lee, H.K.H., Clyde, M.A.: Lossless online bayesian bagging. Journal of Machine
  Learning Research  \textbf{5},  143--151 (2004)

\bibitem{Littlestone:1994}
Littlestone, N., Warmuth, M.K.: The weighted majority algorithm. Inf. Comput.
  \textbf{108}(2),  212--261 (Feb 1994)

\bibitem{Liu:2017}
Liu, N., Zhu, W., Liao, B., Ren, S.: Weighted ensemble with dynamical chunk
  size for imbalanced data streams in nonstationary environment (01 2017).
  \doi{10.2991/iccia-17.2017.60}

\bibitem{Lu:2020}
Lu, Y., Cheung, Y.M., Yan~Tang, Y.: Adaptive chunk-based dynamic weighted
  majority for imbalanced data streams with concept drift. IEEE Transactions on
  Neural Networks and Learning Systems  \textbf{31}(8),  2764--2778 (2020).
  \doi{10.1109/TNNLS.2019.2951814}

\bibitem{Maciel:2015}
Maciel, B.I.F., Santos, S.G.T.C., Barros, R.S.M.: A lightweight concept drift
  detection ensemble. In: 2015 IEEE 27th International Conference on Tools with
  Artificial Intelligence (ICTAI). pp. 1061--1068 (Nov 2015)

\bibitem{Minku:2010}
Minku, L.L., White, A.P., Yao, X.: The impact of diversity on online ensemble
  learning in the presence of concept drift. IEEE Trans. on Knowl. and Data
  Eng.  \textbf{22}(5),  730–742 (may 2010). \doi{10.1109/TKDE.2009.156},
  \url{https://doi.org/10.1109/TKDE.2009.156}

\bibitem{Mohamad:2018}
Mohamad, S., Bouchachia, A., Sayed-Mouchaweh, M.: A bi-criteria active learning
  algorithm for dynamic data streams. IEEE Transactions on Neural Networks and
  Learning Systems  \textbf{29}(1),  74--86 (Jan 2018).
  \doi{10.1109/TNNLS.2016.2614393}

\bibitem{skmultiflow}
Montiel, J., Read, J., Bifet, A., Abdessalem, T.: Scikit-multiflow: A
  multi-output streaming framework. Journal of Machine Learning Research
  \textbf{19}(72), ~1--5 (2018)

\bibitem{Nguyen:2013}
Nguyen, H.L., Ng, W.K., Woon, Y.K.: Concurrent Semi-supervised Learning with
  Active Learning of Data Streams, pp. 113--136. Springer Berlin Heidelberg,
  Berlin, Heidelberg (2013)

\bibitem{Oliveira:2021}
Oliveira, G., Minku, L., Oliveira, A.: Tackling virtual and real concept
  drifts: An adaptive gaussian mixture model (2021)

\bibitem{Oza:2001}
Oza, N.C., Russell, S.: Online bagging and boosting. In: Proceedings of the
  Eighth International Workshop on Artificial Intelligence and Statistics
  (AISTATS'01). p. 105112. Morgan Kaufmann, Key West, USA (2001)

\bibitem{Page:1954}
Page, E.S.: {Continuous Inspection Schemes}. Biometrika  \textbf{41}(1/2),
  100--115 (1954)

\bibitem{Partridge:1997}
Partridge, D., Krzanowski, W.: Software diversity: practical statistics for its
  measurement and exploitation. Information and Software Technology
  \textbf{39}(10),  707 -- 717 (1997)

\bibitem{scikit-learn}
Pedregosa, F., Varoquaux, G., Gramfort, A., Michel, V., Thirion, B., Grisel,
  O., Blondel, M., Prettenhofer, P., Weiss, R., Dubourg, V., Vanderplas, J.,
  Passos, A., Cournapeau, D., Brucher, M., Perrot, M., Duchesnay, E.:
  Scikit-learn: Machine learning in {P}ython. Journal of Machine Learning
  Research  \textbf{12},  2825--2830 (2011)

\bibitem{Ross:2012}
Ross, G.J., Adams, N.M., Tasoulis, D.K., Hand, D.J.: Exponentially weighted
  moving average charts for detecting concept drift. Pattern Recognition
  Letters  \textbf{33}(2),  191 -- 198 (2012)

\bibitem{SantosADOB:2012}
Santos, S., Gonçalves~Jr, P., Silva, G., Barros, R.: Speeding up recovery from
  concept drifts. pp. 179--194 (09 2014)

\bibitem{Schlimmer:1986}
Schlimmer, J.C., Granger, Jr., R.H.: Incremental learning from noisy data.
  Mach. Learn.  \textbf{1}(3),  317--354 (Mar 1986)

\bibitem{Settles:2012}
Settles, B.: Active Learning. Synthesis Lectures on Artificial Intelligence and
  Machine Learning, Morgan {\&} Claypool Publishers (2012).
  \doi{10.2200/S00429ED1V01Y201207AIM018},
  \url{https://doi.org/10.2200/S00429ED1V01Y201207AIM018}

\bibitem{Shan:2018}
Shan, J., Zhang, H., Liu, W., Liu, Q.: Online active learning ensemble
  framework for drifted data streams. IEEE Transactions on Neural Networks and
  Learning Systems  \textbf{30}(2),  486--498 (2019).
  \doi{10.1109/TNNLS.2018.2844332}

\bibitem{Sobolewski:2013}
Sobolewski, P., Wozniak, M.: Concept drift detection and model selection with
  simulated recurrence and ensembles of statistical detectors. Journal of
  Universal Computer Science  \textbf{19}(4),  462--483 (feb 2013)

\bibitem{SouzaChallenges:2020}
Souza, V.M.A., Reis, D.M., Maletzke, A.G., Batista, G.E.A.P.A.: Challenges in
  benchmarking stream learning algorithms with real-world data. Data Mining and
  Knowledge Discovery  \textbf{34},  1805--1858 (2020).
  \doi{10.1007/s10618-020-00698-5}

\bibitem{stapor2021design}
Stapor, K., Ksieniewicz, P., Garc{\'\i}a, S., Wo{\'z}niak, M.: How to design
  the fair experimental classifier evaluation. Applied Soft Computing
  \textbf{104},  107219 (2021)

\bibitem{Street2001}
Street, N., Kim, Y.: A streaming ensemble algorithm (sea) for large-scale
  classification. Proceedings of the 7Th ACM SIGKDD International Conference on
  Knowledge Discovery and Data Mining pp. 377--382 (01 2001)

\bibitem{Street:2001}
Street, W.N., Kim, Y.: A streaming ensemble algorithm (sea) for large-scale
  classification. In: Proceedings of the seventh ACM SIGKDD international
  conference on Knowledge discovery and data mining. pp. 377--382. KDD '01,
  ACM, New York, NY, USA (2001)

\bibitem{Zliobaite:2014}
\v{Z}liobait\.{e}, I., Bifet, A., Pfahringer, B., Holmes, G.: Active learning
  with drifting streaming data. {IEEE} Trans. Neural Netw. Learning Syst.
  \textbf{25}(1),  27--39 (2014)

\bibitem{Wang:2003b}
Wang, H., Fan, W., Yu, P.S., Han, J.: Mining concept-drifting data streams
  using ensemble classifiers. In: Proceedings of the ninth ACM SIGKDD
  international conference on Knowledge discovery and data mining. pp.
  226--235. KDD '03, ACM, New York, NY, USA (2003)

\bibitem{Wang:2015}
Wang, S., Minku, L.L., Yao, X.: Resampling-based ensemble methods for online
  class imbalance learning. {IEEE} Trans. Knowl. Data Eng.  \textbf{27}(5),
  1356--1368 (2015)

\bibitem{Webb:2018}
Webb, G., Hyde, R., Cao, H., Nguyen, H.L., Petitjean, F.: Characterizing
  concept drift. Data Mining and Knowledge Discovery  \textbf{30} (07 2016).
  \doi{10.1007/s10618-015-0448-4}

\bibitem{Widmer:1993}
Widmer, G., Kubat, M.: Effective learning in dynamic environments by explicit
  context tracking. In: Brazdil, P. (ed.) Machine Learning: ECML-93, Lecture
  Notes in Computer Science, vol.~667, pp. 227--243. Springer Berlin Heidelberg
  (1993)

\bibitem{Widmer:1996}
Widmer, G., Kubat, M.: Learning in the presence of concept drift and hidden
  contexts. Mach. Learn.  \textbf{23}(1),  69--101 (Apr 1996)

\bibitem{Wozniak:2013}
Wozniak, M., Kasprzak, A., Cal, P.: Application of combined classifiers to data
  stream classification. In: Proceedings of the 10th International Conference
  on Flexible Query Answering Systems FQAS 2013. pp. 13--23. LNCS,
  Springer-Verlag, Berlin, Heidelberg (2013)

\bibitem{Xu:2016}
Xu, W., Zhao, F., Lu, Z.: Active learning over evolving data streams using
  paired ensemble framework. In: 2016 Eighth International Conference on
  Advanced Computational Intelligence (ICACI). pp. 180--185. IEEE (2016)

\bibitem{Zgraja:2018}
Zgraja, J., Gama, J., Wozniak, M.: Active learning by clustering for drifted
  data stream classification. In: {ECML} {PKDD} 2018 Workshops - {DMLE} 2018
  and IoTStream 2018, Dublin, Ireland, September 10-14, 2018, Revised Selected
  Papers. pp. 80--90 (2018). \doi{10.1007/978-3-030-14880-5\_7},
  \url{https://doi.org/10.1007/978-3-030-14880-5\_7}

\bibitem{zyblewski2020combination}
Zyblewski, P., Ksieniewicz, P., Wo{\'z}niak, M.: Combination of active and
  random labeling strategy in the non-stationary data stream classification.
  In: International Conference on Artificial Intelligence and Soft Computing.
  pp. 576--585. Springer (2020)

\bibitem{Lapinski:2018}
Łapiński, A., Krawczyk, B., Ksicnicwicz, P., Woźniak, M.: An empirical
  insight into concept drift detectors ensemble strategies. In: 2018 IEEE
  Congress on Evolutionary Computation (CEC). pp.~1--8 (2018).
  \doi{10.1109/CEC.2018.8477962}

\end{thebibliography}

\end{document}